%% file: main.tex
\documentclass{article}

\usepackage{PRIMEarxiv}

\usepackage[numbers]{natbib}
\usepackage[utf8]{inputenc} 
\usepackage[T1]{fontenc}    
\usepackage{hyperref}       
\usepackage{booktabs}       
\usepackage{amsfonts}       
\usepackage{nicefrac}       
\usepackage{microtype}      
\usepackage{xcolor}         
\usepackage{enumitem}
\usepackage{graphicx}
\usepackage{subcaption}
\usepackage{amsmath}
\usepackage{pifont}
\usepackage{wrapfig}
\usepackage{array}
\usepackage{amsthm}

\newcommand{\cmark}{\ding{51}}
\newcommand{\xmark}{\ding{55}}
\newcommand{\D}{\mathcal{D}}

\title{Tabular Few-Shot Generalization Across Heterogeneous Feature Spaces}

\author{Max Zhu$^{1*}$ \quad Katarzyna Kobalczyk$^{1*}$ \quad Andrija Petrovic$^{2}$ \quad Mladen Nikolic$^{3}$\\
     \quad \textbf{Mihaela van der Schaar}$^{1}$ \quad \textbf{Boris Delibasic}$^{2}$ \quad \textbf{Petro Lio}$^{1}$ \vspace{0.5em}\\ 
     $^{1}$University of Cambridge \quad $^{2, 3}$University of Belgrade \vspace{0.5em}\\
     \texttt{\{mz406, knk25, mv472, pl219\}@cam.ac.uk}\\
     \texttt{\{andrija.petrovic, boris.delibasic\}@fon.bg.ac.rs}\\
     \texttt{\{mladen.nikolic\}@matf.bg.ac.rs}
}

\begin{document}
\maketitle

\begin{abstract}
    Despite the prevalence of tabular datasets, few-shot learning remains under-explored within this domain. Existing few-shot methods are not directly applicable to tabular datasets due to varying column relationships, meanings, and permutational invariance. To address these challenges, we propose FLAT—a novel approach to tabular few-shot learning, encompassing knowledge sharing between datasets with \textit{heterogeneous feature spaces}. Utilizing an encoder inspired by Dataset2Vec, FLAT learns low-dimensional embeddings of datasets and their individual columns, which facilitate knowledge transfer and generalization to previously unseen datasets. A decoder network parametrizes the predictive target network, implemented as a Graph Attention Network, to accommodate the heterogeneous nature of tabular datasets. Experiments on a diverse collection of 118 UCI datasets demonstrate FLAT’s successful generalization to new tabular datasets and a considerable improvement over the baselines.
    
\end{abstract}

\section{Introduction}
Few-shot learning is a machine learning paradigm in which models are trained to make accurate predictions with only a few labeled examples, often leveraging prior knowledge obtained from training on a collection of related tasks ~\citep{song2022comprehensive, wang2020generalizing}. While few-shot learning techniques have been extensively studied in computer vision (CV) and natural language processing (NLP)~\citep{oh2020boil, kang2021relational, perez2021true}, tabular data has received little attention, despite its importance in many practical applications, including finance~\citep{cao2022ai}, healthcare~\citep{shailaja2018machine}, and social sciences~\citep{molina2019machine}. However, such applications often suffer from limited labeled data due to its rarity or high labeling costs. For example, in finance~\citep{bhatore2020machine}, determining credit risk requires considerable effort in data labeling, and in healthcare~\citep{schaefer2020use}, rare diseases may not have enough samples to train a robust model from scratch. 

Few-shot learning on tabular data has been explored on a very limited scale—mostly assuming that the training and target datasets share the same feature space \citep{hegselmann2022tabllm, nam2023stunt}. Generalizing tabular few-shot learning across heterogeneous tabular datasets poses unique challenges. Firstly, columns of such datasets have no intrinsic meaning transferable between different datasets; they are assigned meaning strictly in the context of their relationships to other columns within the same dataset. This is in contrast to natural language data, where each word always corresponds to a fixed set of meanings. Secondly, tabular datasets exhibit varying column-label relationships; tabular datasets can follow different distributions and there is no obvious way in which different datasets can relate to each other. Finally, tabular data exhibits permutational invariance with respect to the column order, unlike image and text data, where meaning depends on the order of words or pixels. For these reasons, existing methods developed for CV and NLP cannot be directly applied to tabular datasets.

To address these challenges, we propose FLAT—tabular \textbf{F}ew-shot \textbf{L}earning with graph \textbf{AT}tention networks. FLAT is formulated within the meta-learning paradigm of \citet{vinyals2016matching}. FLAT consists of a meta network, which given a small few-shot sample, generates weights for the target network. The meta network employs an encoder-decoder architecture. The encoder, inspired by Dataset2Vec \citep{jomaa2021dataset2vec}, embeds datasets and their individual columns into low-dimensional shared subspaces and the decoder generates the weights for the target network. The target network, a Graph Attention Network (GAT) \citep{velickovic2017graph}, operates with these embeddings to perform inference on unlabeled instances. This solves the challenges outlined in the paragraph above: a) the dataset encoder and the target GAT network make FLAT permutation-invariant; b) the column embeddings combined with GAT enable dynamic assignment of meaning and relations to features; c) the shared embedding space of datasets facilitates meta-learning across datasets. We verify the effectiveness of FLAT on 118 classification datasets from the UCI repository~\citep{duaUCI} and show that FLAT considerably outperforms existing methods, including tasks with highly imbalanced classes in the target variable.

{\bf Contributions} 1)~We introduce column embeddings and permutation invariant dataset embeddings to facilitate knowledge transfer within shared low-dimensional subspaces across datasets with varying sets of features. 2)~We design meta and target networks suited for heterogeneous tabular datasets to exploit inter-column relationships and identify structural similarities between training and test datasets, thereby enabling tabular few-shot generalization. 3)~We compose these elements into a novel few-shot learning method that generalizes over tabular datasets with varying sets of columns, as demonstrated by the experimental evaluation. 

\section{Related work} \label{sec:Related Work}
\textbf{Meta- and few-shot learning}
Few-shot learning aims to train models capable of adapting to new tasks with minimal labeled examples. This can be achieved through meta-learning on a variety of related tasks to obtain prior knowledge that can be leveraged to solve new tasks~\citep{ wang2020generalizing, jadon2020overview}. Notable approaches to few-shot meta-learning include learning a distance metric ~\citep{vinyals2016matching, prototypical}, parameter initializations~\citep{raghu2019rapid, li2017meta}, parameter generators~\citep{li2019lgm, hypernetworks}, or learning the learning algorithms~\citep{metalstm}.
Our few-shot meta-learning model is inspired by LGM-Net \citep{li2019lgm}, which generates task embeddings from a sample of data, and conditioned on these embeddings, samples weights for the target matching network~\citep{vinyals2016matching} that solves the target task. Through dynamic weight generation, the model adjusts its behavior to best suit the input task. While we keep the base idea, we adapt it to tabular data.

\textbf{Attention based models for tabular data}
Recently, attention-based models have achieved state-of-the-art performance in tabular deep learning. Among them, TabNet \citep{arik2021tabnet} and FT-Transformer \citep{FTtransformer} are two notable examples. By utilizing sequential attention to select the most important features, TabNet improves interpretability and learning efficiency. FT-Transformer \citep{FTtransformer} is an adaptation of the Transformer architecture \citep{AttentionNeed} to the tabular domain and can be seen as an evolution of TabTransformer \citep{huang2020tabtransformer}. The model transforms all features into tokens and runs a stack of transformer layers over the tokens. Inspired by the success of attention-based architectures, we implement the target network as a GAT~\citep{velickovic2017graph}—a type of graph neural network \citep{GNN-review} employing the attention mechanism to pass information between the nodes of a graph leading to improved performance compared to simpler baselines like the Graph Convolutional Network \citep{GCN}.

\begin{figure*}[h]
    \centering
    \includegraphics[width=0.8\textwidth]{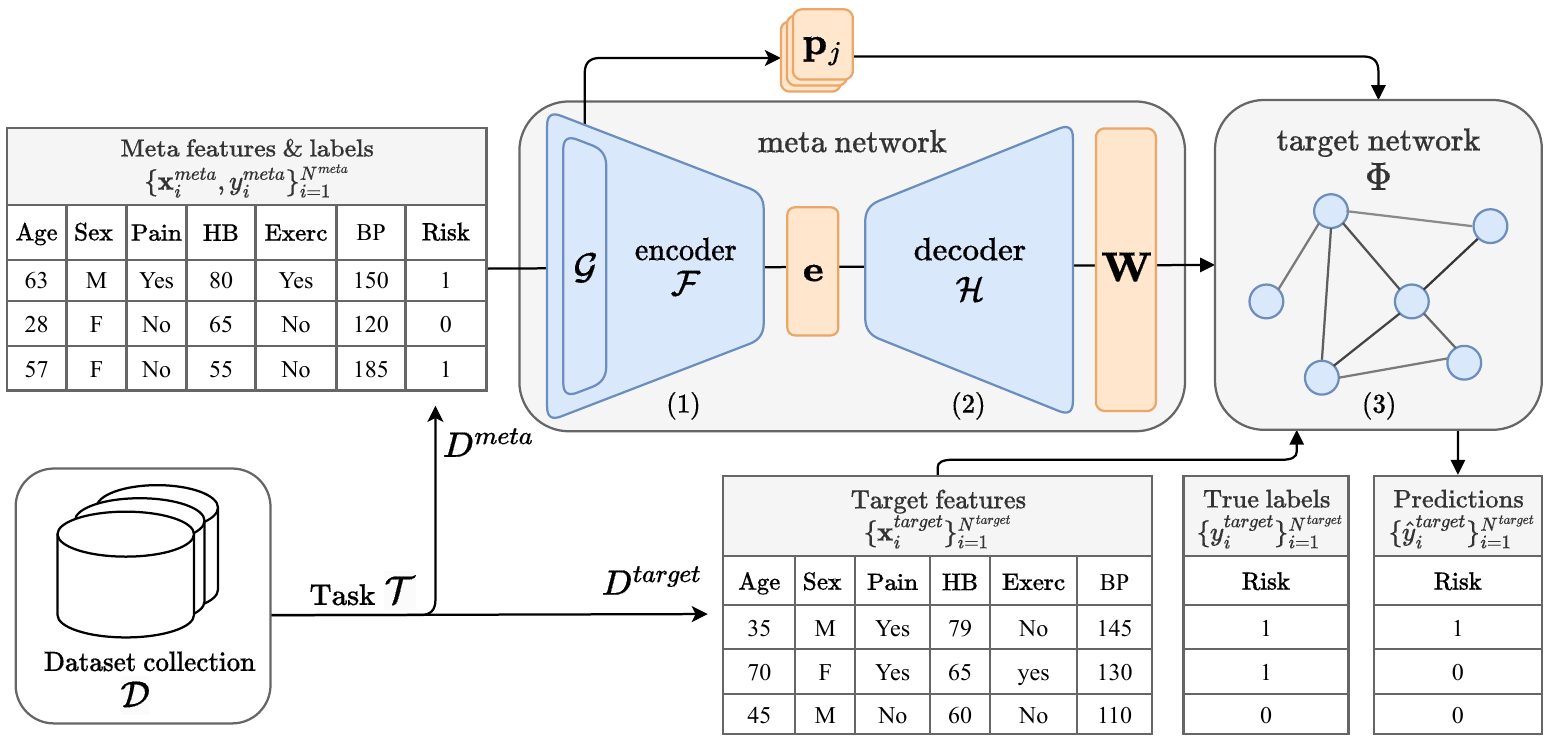}
    \caption{Overview of the FLAT architecture, highlighting its three key components: (1—dataset encoder $\mathcal{F}$ with the column encoder $\mathcal{G}$, (2)—weight generating decoder network $\mathcal{H}$ and (3)—the target GAT network $\Phi$. (1) and (2) together form the meta network.}
    \label{fig:FLAT}
\end{figure*}

\textbf{Tabular few-shot learning}
Most research on few-shot learning focuses on NLP and CV tasks. While a small subset of approaches explicitly tackles tabular few-shot learning, many of them exhibit notable limitations:

TabLLM \citep{hegselmann2022tabllm} fine-tunes large language models (LLMs) on tabular datasets serialized into natural language. The LLM uses its semantic knowledge to improve classification accuracy. TabLLM requires access to meaningful names of the predictors, which may not be available (e.g. when working with anonymized datasets). Moreover, black-box LLMs suffer from limited interpretability and are susceptible to undesirable biases \citep{bias}. 

STUNT \citep{nam2023stunt} meta-learns generalizable knowledge from few-shot tasks, self-generated from an unlabeled set of examples. To generate the meta-tasks, STUNT requires an additional unlabeled training dataset of a considerable size that shares the same feature space as the test dataset. Yet, such data may be unavailable or difficult to obtain. 

\citet{Iwata2020} propose a heterogeneous meta-learning method based on Deep Sets \citep{zaheer2017deep} operators. Their method learns separate latent representations of each attribute and response column, which together with the unlabeled features are passed as inputs to the predictive network. This simple architecture has proven successful on regression tasks, yet their evaluation on classification tasks is limited to small artificial binary classification tasks. Moreover, while Deep Sets are easy to implement, processing each column of a dataset individually can hinder relational reasoning and feature interactions \citep{wagstaffDeepSets}, thus limiting the performance gains.

TabPFN \citep{hollmann2023tabpfn} is a transformer-based prior-data fitted network that approximates Bayesian inference by training on synthetic data generated from prior distributions mimicking real-world data generation mechanisms. TabPFN is designed to make fast and accurate predictions on a single ``small'' dataset. However, it is not intended for transferring knowledge between existing real-world datasets and a downstream dataset containing just a few labeled samples. Moreover, its input size is limited to its training size ($\leq 1000$ labeled samples, $\leq 100$ features, $\leq$ 10 classes). 

In contrast to previous works, FLAT does not require semantically meaningful column names or a large number of unlabeled samples. FLAT successfully captures structural relationships between the features and transfers knowledge between real-world datasets of varying feature spaces, outperforming all existing baselines on few-shot classification tasks. In addition, FLAT offers a higher degree of interpretability through the visualization of attention weights and dataset embeddings.

\section{FLAT: Tabular Few-Shot Learning with Graph Attention Networks}\label{sec:Methodology}

In this section, we clearly define the problem FLAT aims to solve, followed by a description of the model architecture and the training procedure. The model overview of FLAT is presented in detail in Fig~\ref{fig:FLAT}.

\subsection{Problem definition}
A task $\mathcal{T}$ is defined by a small meta dataset $D^{meta} = \{(\mathbf{x}_i^{meta}, y_{i}^{meta})\}_{i=1}^{N^{meta}}$ and a target dataset $D^{target} = \{(\mathbf{x}_i^{target}, y_i^{target})\}_{i=1}^{N^{target}}$, where $\mathbf{x}_i^{meta},\mathbf{x}_i^{target}\in\mathbb{R}^{N^{col}}$ are feature vectors of size $N^{col}$, $y_i^{meta}, y_i^{target}\in\mathcal{Y}$ are the corresponding labels, and $N^{meta}$ and $N^{target}$ are the number of samples in the meta and target datasets respectively. The number of columns $N^{col}$ in each task can vary between the tasks. We assume that for a single task $\mathcal{T}$, $D^{meta}$ and $D^{target}$ follow the same data distribution. During testing, $D^{meta}$ is labeled while only the features of $D^{target}$, i.e. $\mathbf{x}^{target}$, are known. Our goal is to train a model $\mathcal{M}$ to predict unknown labels $y^{target}$ using $D^{meta}$ and $\mathbf{x}^{target}$. $\mathcal{M}$ should generalize well to unseen tasks generated from different data distributions. In this paper, we mainly focus on binary classification tasks, where $\mathcal{Y} = \{0, 1\}$. We also demonstrate FLAT's performance on 3-class classification tasks.

\subsection{FLAT}


{\bf Model structure} Our model can be decomposed into three main parts: 
(1)—the permutation-invariant encoders, $\mathcal{F}$ and $\mathcal{G}$, which produce dataset embeddings $\mathbf{e}$ and column embeddings $\mathbf{p}_j$, 
(2)—the decoder $\mathcal{H}$, which generates the weights $\mathbf{W}$ based on the dataset embedding, and (3)—the target network $\Phi$, a fully connected GAT. The first two elements form the meta network, which parametrizes the target network. 

The encoder maps a dataset into a shared embedding space of all datasets, $\mathbf{e} \in \mathbb{R}^{d_e}$. The embeddings capture important dataset characteristics, such that similar datasets are close to one another in the embedding space. Similarly, individual columns are mapped into a column embedding space, $\mathbf{p}_j \in \mathbb{R}^{d_c}$ for $j \in \left[N^{col}\right]$. $d_e$ and $d_c$ are the dimensions of the dataset and column embeddings, respectively. The target network is conditioned on these embeddings, enabling it to adjust its behavior to a particular dataset. By mapping all datasets into a fixed-dimension latent space, our model can process and relate together different tabular datasets, even with non-overlapping sets of features.

{\bf Model training and testing} We let $\D_{train}$ and $\D_{test}$ denote collections of datasets used for training and testing, respectively. In each training iteration, we first sample a dataset from $\D_{train}$ and extract from it a small subsample forming the meta-task $\mathcal{T}=(D^{meta}, D^{target})$. The meta network encodes $D^{meta}$ and generates target network parameters. The target network performs inference on the features of $D^{target}$ and generates predictions $\hat{\mathbf{y}}^{target}$. During training, a binary cross-entropy loss is computed between the predictions $\hat{\mathbf{y}}^{target}$ and the ground truth labels $\mathbf{y}^{target}$. Weights are then updated with backpropagation. Once trained, FLAT performs inference on tasks generated from unseen datasets from $\D_{test}$, following the same procedure as during training. 

\subsubsection{The meta network}

At the core of our meta network lies the \textbf{dataset encoder} $\mathcal{F}$, which extracts important characteristics of a dataset for downstream classification. $\mathcal{F}$ takes in a tabular dataset of any size and outputs a permutation invariant embedding vector of fixed dimension. We base $\mathcal{F}$ on Dataset2Vec \citep{jomaa2021dataset2vec}. Our variant is defined as:
\begin{equation}\label{eq:d2v}
\small
\mathbf{e} = f_3\left(\frac{1}{N^{col}} \sum\limits_{j=1}^{N^{col}}    f_2\left(\frac{1}{N^{meta}} \sum\limits_{i=1}^{N^{meta}}f_1(x_{i, j}^{meta}, y_{i}^{meta})\right)\right),
\end{equation}
where $f_1$, $f_2$ and $f_3$ are MLP blocks, and $N^{col}$ is the number of columns. The inner sum spans rows, and the outer sum spans feature columns, ensuring $\mathcal{F}$ is permutation-invariant across rows and columns. Unlike the original Dataset2Vec's contrastive loss, we directly train $\mathcal{F}$ as part of the end-to-end training scheme with no explicit constraints on $\mathbf{e}$.  

The \textbf{column encoder $\mathcal{G}$} generates column embeddings $\mathbf{p}_j$ as in equaion (\ref{eq:embeddings}). It applies an MLP $g$ to the first stage of $\mathcal{F}$ after summing over rows, capturing the relation between a single column and labels.
\begin{equation}\label{eq:embeddings}
\small
\mathbf{p}_j = g\left(\frac{1}{N^{meta}} \sum\limits_{i=1}^{N^{meta}}f_1(x_{i, j}^{meta}, y_{i}^{meta}) \right)
\end{equation}
The \textbf{weight decoder $\mathcal{H}$} is a set of $L$ MLPs \{$h_1\ldots,h_L$\} where $L$ is the number of layers in the target network. For $l=1\ldots,L-1$, $h_l$ generates GAT weights from a dataset embedding $\mathbf{e}$ :
\begin{gather}
\small
\left[\omega_a^l,\omega_b^l,\omega_W^l\right] = h_l(\mathbf{e}), \\
\mathbf{a}^l = \theta_a \frac{\omega_a^l}{\|\omega_a^l\|}, \quad 
\mathbf{b}^l = \theta_b \frac{\omega_b^l}{\|\omega_b^l\|},\quad
\mathbf{W}^l = \theta_w \frac{\omega_W^l}{\|\omega_W^l\|},  
\end{gather}
where $\mathbf{a}^l$ and $\mathbf{b}^l$ are vectors of attention weights and biases and $\mathbf{W}^l$ is the matrix of feature transformation weights.
For $l=L$, corresponding to the final linear classifier, only $\mathbf{W}^{L}$ is generated.
Like LGM-Net, we apply L2-normalization to the generated weights \citep{WeightNorm}, yet we let $\theta$ be learnable and do not use weight sampling or reparameterization.

\subsubsection{The target network}
We opt for a GAT as the target network, $\Phi$ (presented without bias terms $\textbf{b}^l$ for brevity). $\Phi$ consists of several GAT layers, followed by a linear classification layer. The attention coefficients $\alpha_{jk}$ and the hidden states of the next GAT layer $\mathbf{h}_j^{l+1}$ are computed as:
\begin{gather}\label{eq:GAT}
\small
\alpha_{jk} = \frac{
\exp\left(\text{LReLU} \left({\mathbf{a}^l}^{\top}
[\mathbf{W}^l\mathbf{h}^l_j \, \Vert \, \mathbf{W}^l\mathbf{h}^l_k]
\right)\right)}
{\sum_{r \in \mathcal{N}_j}
\exp\left(\text{LReLU}\left({\mathbf{a}^l}^{\top}
[\mathbf{W}^l\mathbf{h}^l_j \, \Vert \, \mathbf{W}^l\mathbf{h}^l_r]
\right)\right)}, \\
\mathbf{h}^{l+1}_j = \sum_{k \in \mathcal{N}_j} \alpha_{jk}\mathbf{W}^l\mathbf{h}^l_{k},
\end{gather}
where $\mathbf{h}^l_j$ is the embedding of node $j$ computed by layer $l$, $\mathcal{N}_j$ are neighboring nodes of $j$ including itself, $\mathbf{W}^l$ and $\mathbf{a}^l$ are parameters provided by the weight generating network, $\text{LReLU}$ is the Leaky ReLU activation, and $\Vert$ denotes concatenation. The first layer node vectors $\mathbf{h}_j^0 = \left[\mathbf{p}_j || x_j\right]$, $j\in\left[N^{col}\right]$, are concatenations of column embeddings and its feature value.
Each GAT layer operates on a fully connected graph where every node corresponds to one feature. The attention coefficients and hidden states of the GAT are computed independently for each row $i \in \left[N^{target}\right]$ of the target dataset $D^{target}$, while the parameters $\mathbf{a}^l, \mathbf{b}^l, \mathbf{W}^l$ are shared across all rows. 

To obtain predictions, the final GAT hidden layer node representations $\mathbf{h}_j^{L-1}$ are averaged and passed to a linear classifier with 2 output heads
\begin{equation}
\small
p(\hat{y}^{target}) = \text{softmax}\left(\mathbf{W}^L  \left( \frac{1}{N^{col}}\sum_j{\mathbf{h}_j^{L-1}} \right)\right).
\end{equation}

GATs are a suitable architecture since they can process graphs of any size, corresponding to datasets with any number of features. GATs use the same weights for each node, and our graph is fully connected, meaning that $\Phi$ is fully permutation invariant while using fewer parameters than an equivalent-size transformer. However, the target network must be invariant to column \emph{order} while \emph{identifying} which columns in $D^{meta}$ correspond to $D^{target}$. Concatenating column embeddings to feature values allows the network to identify and interpret different features in different ways. Furthermore, when combined with the fully connected attention mechanism, column embeddings allow the GAT to consider interactions between features.

\subsubsection{FLATadapt}\label{sec:flat-adapt}
As shown by the experimental evaluation in section~\ref{sec:exp-evaluation}, FLAT is able to bring competitive performance against the baselines. 
We also present a further extension—FLATadapt.
FLATadapt takes a \textit{pre-trained} FLAT model and adapts the dataset embeddings $\textbf{e}$, and column embeddings, $\textbf{p}_j$ with a few steps of gradient descent on the features and labels of $D^{meta}$, but only at inference time. All model weights remain unchanged. This method only changes \textit{how to perform inference on an already-trained FLAT model}, avoiding additional complexity during training (see Appendix \ref{Appendix:details} for implementation details). In section \ref{sec:generalist-exp}, we demonstrate that the extra adaptation step can increase performance at the cost of longer inference time.

\section{Experimental evaluation}
\label{sec:exp-evaluation} 
In this section, we validate the effectiveness of our method in few-shot tabular learning using a collection of 118 tabular classification datasets from the UCI Machine Learning Repository \citep{duaUCI}.

\textbf{Experimental setup} First, to increase the number and variety of binary classification tasks, the dependent variables of datasets with more than two prediction classes (65 of 118) were binarized by setting the most common class as positive and all other classes as negative (one-vs-all). FLAT models are trained and tested using an $N$-fold evaluation procedure. We split the collection of all datasets into $N$ folds. Each fold is then used once as the testing collection $\D_{test}$, while the remaining $N-1$ folds form $\D_{train}$. To generate a task during training or testing, a dataset is chosen uniformly at random from the relevant collection ($\mathcal{D}_{train}$ or $\mathcal{D}_{test}$). Then, $N^{meta} + N^{target}$ rows are sampled to form $D^{meta}$ and $D^{target}$. Feature columns are standardized to mean 0 and variance 1. During training, as a form of data augmentation, we randomly subsample varying numbers of feature columns for both $D^{meta}$ and $D^{target}$, allowing the model to be exposed to a wider range and difficulty of tasks.
FLAT results are averaged over multiple random seeds.

\textbf{Imbalanced few-shot learning} Our setup differs from the conventional $K$-shot learning, where meta datasets contain an equal number of examples per class. Unless otherwise stated, we employ a randomized sampling procedure. The number of positive examples in $D^{meta}$ and $D^{target}$ are sampled from a binomial distribution with success probability $p=0.5$. For a fair comparison against fully supervised learning algorithms, we require that $D^{meta}$ contains at least one example of each class (except when $N^{meta} = 1$).  This approach simulates a more realistic scenario in which task datasets may often have imbalanced classes. For example, rare diseases may have a prevalence rate of only 0.1\%. A conventional $5$-shot learning approach would require around 5,000 records in order to construct a meta dataset with 5 positive and 5 negative samples. The standard $K$-shot and binomial sampling approaches are compared in Appendix \ref{Appendix-classic-fewshot}. 

\textbf{Baselines} We evaluate our approach against:
\begin{itemize}[label={--}, itemsep=0pt, topsep=0pt]
    \item standard supervised learning models: logistic regression (LR), k-nearest neighbors (KNN), support vector classifier (SVC), random forest classifier (RForest),  CatBoost \citep{CatBoost},
    \item supervised deep-learning models for tabular data: TabNet \citep{arik2021tabnet}, FT-Transformer (FTT) \citep{FTtransformer},
    \item semi-supervised meta-learning model for tabular data—STUNT \citep{nam2023stunt},
    \item prior-data fitted supervised classifier for tabular data—TabPF \citep{hollmann2023tabpfn},
    \item few-shot meta-learning model for tabular data of \cite{Iwata2020} (Iwata).
\end{itemize}
We do not compare against TabLLM since our setup does not assume access to semantically meaningful columns. Iwata is meta-trained and tested using the same $N$-fold evaluation procedure as FLAT. All remaining baselines require a training dataset with the same feature space as the test dataset. By our assumption, the only labeled samples with the same feature space are those found in $D^{meta}$. Therefore, all baselines (except Iwata) are fitted on $D^{meta}$, and their performance is evaluated on $D^{target}$ independently for each task. For STUNT, we run the pre-training procedure on $\{\mathbf{x}_i^{meta}\}_{i=1}^{N^{meta}}$ and use $\{y^{meta}_i\}_{i=1}^{N^{meta}}$ as prototypes.


\textbf{Validation} As our setup does not assume access to labeled samples beyond  $D^{meta}$ that could be used for hyperparameter tuning, we use a validation procedure that identifies a global set of hyperparameters, leading to good generalization performance across multiple datasets instead of tuning them for each dataset separately. To achieve this, a collection of validation tasks, $\D_{val}$, is generated by randomly selecting 25\% of all 118 UCI datasets and subsampling 25\% of rows, ensuring that there is no overlap between validation and testing rows. Hyperparameters for all models were selected by maximizing the accuracy on tasks sampled from $\D_{val}$ and are fixed throughout all testing runs. 

Full details of training and hyperparameter tuning for all models are given in Appendix~\ref{Appendix:details} and \ref{Appendix-exp-details}. 

\begin{table*}
    \centering
    \caption{Accuracy (\%) of FLAT vs. the baselines averaged over all testing folds of the medical datasets. $N^{meta}$ labeled meta examples are presented to each model at test time. The best model and those within its error range are highlighted in bold.}
    \label{tab:med-results}
    {\small \include{figures/medical_results}}
\end{table*}

\begin{figure*}
    \centering
    \includegraphics[width=0.4\linewidth]{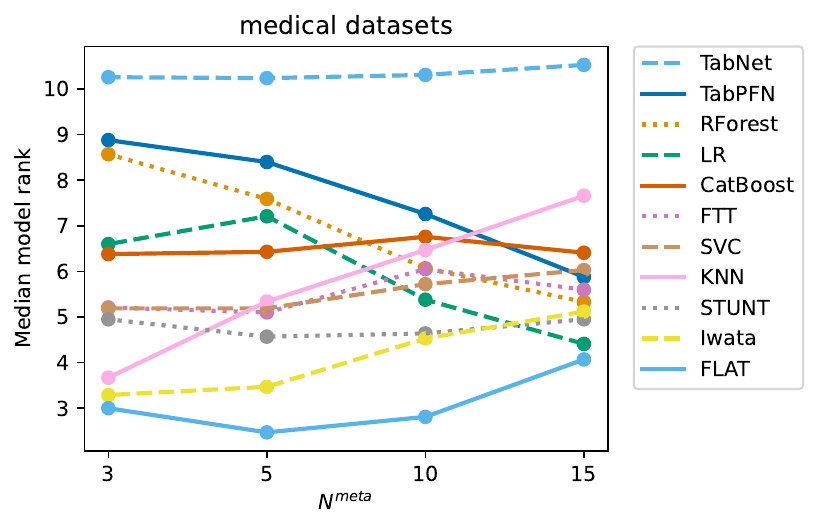}
    \includegraphics[width=0.4\linewidth]{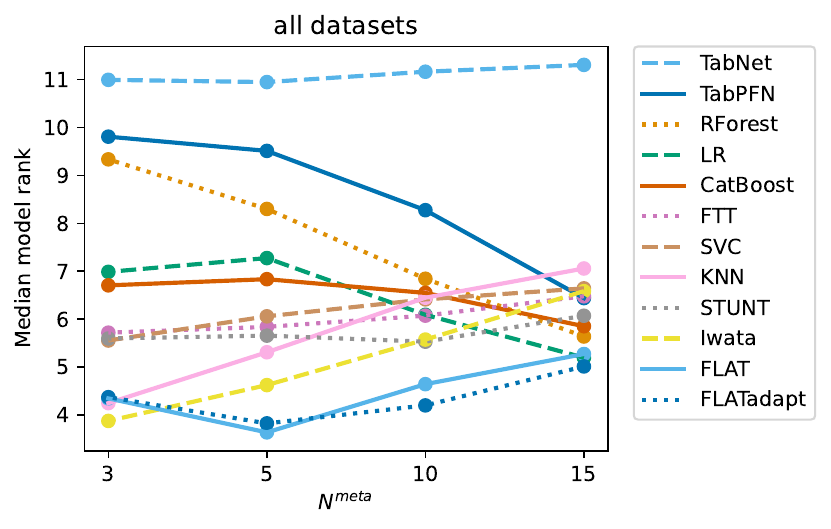}
    \caption{Median model ranks based on accuracy over 29 medical datasets (Left) and all 118 UCI datasets (Right)}
    \label{fig:ranks}
\end{figure*}

\begin{table*}
  \centering
  \caption{Test accuracy (\%) for all datasets. The right column shows the time to run 200 steps of inference at 15 meta and target samples with 20 features. Datasets that are too small to sample from are omitted. The best model and those within its error range are highlighted in bold.}
  {\small\include{figures/all_results}}
  \label{tab:GenAll}
\end{table*}

\begin{figure*}[h]
    \centering
    \includegraphics[width=0.9\textwidth]{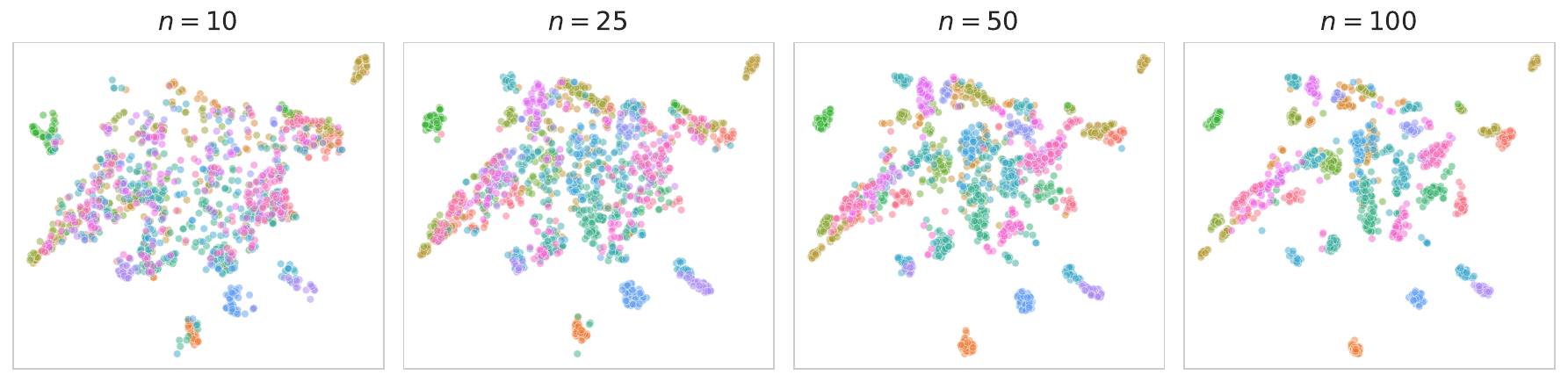}
    \caption{t-SNE plot of the medical datasets embeddings $\mathbf{e}$. Plots generated for increasing number of meta samples $N^{meta} = \min \left(n, \frac{1}{2}N_D^{row}\right)$ for $n \in [10, 25, 50, 100]$, where  $N_D^{row}$ is the total number of rows of the dataset $D$. The embeddings are generated for tasks coming from both $\D_{train}$ and $\D_{test}$. Increasing the number of meta rows reveals the capability of FLAT to cluster together tasks coming from the same datasets.}
    \label{fig:medical-embeddings}
\end{figure*}

\subsection{Illustrative example: medical datasets}\label{sec:medical-example}

The fundamental principle of meta-learning lies in the assumption that different tasks share a certain degree of common knowledge among them. Accordingly, datasets from a single domain represent a promising avenue for successful knowledge transfer. Below, we demonstrate how FLAT can be applied to a subset of 29 UCI datasets associated with medical diagnosis. We selected this subset for illustrative purposes, as a smaller subset of datasets from a known domain allows for easier model interpretation (see Appendix~\ref{sec:Appendix-interpretability}).

As shown in Table~\ref{tab:med-results}, FLAT significantly improves upon the baselines at few-shot tabular classification, with an increase in average accuracy by up to 2pp over the best baseline. FLAT also ranks higher than all baselines for all $N^{meta}$ (Fig.~\ref{fig:ranks}). Detailed results are available in Appendix Fig.~\ref{fig:med-results-bydataset}. Another advantage of pre-trained models like FLAT and Iwata is that they can generate meaningful predictions when the meta dataset contains only a single class. At $N^{meta}=1$, FLAT achieves an average accuracy of 59.7\%, which is a significant improvement over the expected 50\% accuracy for random guessing. The accuracy of FLAT increases with the number of meta samples, yet the relative advantage of FLAT over standard supervised models decreases as $N^{meta}$ increases. This is aligned with FLAT's intended design as a few-shot learner; for a larger number of labeled samples, ``many-shot'' learners become more competitive. 

We demonstrate model interpretability by visualizing the dataset embeddings $\mathbf{e}$ (Fig.~\ref{fig:medical-embeddings}). To reveal the underlying clustering pattern, we sample an increasing number of meta samples to reduce variance in the generated embeddings. As $N^{meta}$ increases, FLAT produces embeddings that form clear clusters in the embedding space. This illustrates that the task encoder learns highly expressive embeddings, allowing the weight-generating network to produce parameters for the target network tailored to each dataset and that t-SNE visualizations are useful in determining which datasets the model considers similar. An additional figure with cluster centroids annotated by the corresponding datasets can be found in Appendix~\ref{sec:Appendix-interpretability}.

\subsection{Training a generalist few-shot 
learner}\label{sec:generalist-exp}

In this section, we use all 118 UCI datasets for training and testing to demonstrate that FLAT can improve few-shot prediction accuracy on datasets spanning multiple domains. We also show how FLATadapt can further improve model performance. 
Results presented in Table \ref{tab:GenAll} show that, on average, FLAT is able to outperform the baselines at $N^{meta}=3, 5, 10$ while matching the baselines at $N^{meta}=15$. FLATadapt consistently improves upon FLAT and exceeds all the baselines by up to  2.33pp. Similarly to the previous example, FLAT(adapt) demonstrates a more substantial performance boost over baselines for smaller $N^{meta}$. 
Additionally, Table \ref{tab:GenAll} displays the time for 200 inferences on tasks with 15 rows and 20 columns. FLAT shows a fast inference time comparable to simple baselines like LR or KNN, while FT-Transformer and TabNet are significantly slower as they need to be re-fitted to each task's meta dataset, which is computationally expensive. FLATadapt is slower than FLAT as it requires a few additional steps of gradient descent during inference. A more detailed comparison of inference time vs. the number of columns is given in Appendix \ref{Appendix-time}.


\subsection{Additional experiments}

\textbf{Multi-class classifcation}
To demonstrate FLAT's applicability to multi-class datasets, we conduct additional experiments on 3-class classification tasks. We select datasets with at least 3 classes (65 in total) and modify the target network to output 3 logits instead~of~2. We train and test FLAT models using the 4-fold evaluation procedure without additional hyperparameter tuning. Table~\ref{tab:multiclass}, in the Appendix shows that FLAT outperforms all baselines at $N^{meta} = 3, 5, 10$ and remains slightly behind at $N^{meta}=15$. FLATadapt improves on FLAT by up to +1.25pp, resulting in the highest average accuracy at $N^{meta} = 3, 5, 10$ and is within the error of the best baselines at $N^{meta} = 15$. 

\begin{wrapfigure}{r}{0.4\textwidth}
    \centering
    \includegraphics[width=\linewidth]{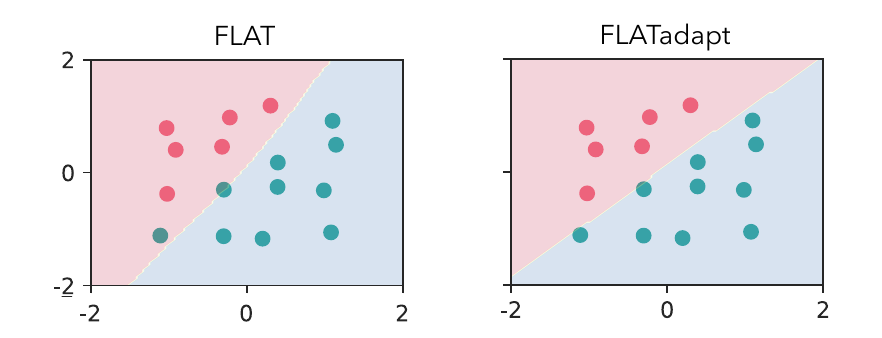}
    \caption{Decision boundaries of a FLAT and FLATadapt on synthetic data. Meta data points are shown as dots. Red is 1, blue is 0. FLAT is misaligned near the boundary which is corrected by FLATadapt.}
    \label{fig:visualize_preds}
\end{wrapfigure}

\textbf{FLATadapt} We visualize the impact of FLATadapt compared to FLAT. 2-D synthetic data (corresponding to 2 columns) is input to a model pre-trained on the UCI datasets. The meta dataset is a perturbed $4 \times 4$ grid with label 1 if $x_1 > x_2$. We plot meta data points and the learned decision boundary in Fig.~\ref{fig:visualize_preds}. FLAT creates a decision boundary that misclassifies two points from the meta dataset. FLATadapt shifts the decision boundary closer to the true boundary, $y=x$, resulting in the correct classification of previously misclassified points.

\textbf{Imbalanced meta datasets} The main body of this paper uses meta datasets that have binomially distributed positive and negative samples. In Appendix \ref{Appendix-classic-fewshot}, we investigate the performance of FLAT depending on how balanced $D^{meta}$ is. FLAT greatly outperforms baselines for imbalanced $D^{meta}$ and is within the error of the best baseline when the $D^{meta}$ is perfectly balanced. 

\textbf{Single sample predictions}  FLAT is able to make predictions with only a single labeled sample, whereas standard supervised models typically require at least one example from each class to perform inference. In Appendix \ref{sec:Appendix-1-shot}, we visualize FLAT's decision boundaries when $N^{meta}=1$ and argue that FLAT essentially learns prior knowledge on how ``close'' a target sample should be to the meta sample in order to be assigned the same class.

\textbf{When does FLAT result in large performance gains?} Training FLAT on all UCI datasets resulted in slightly lower performance gains compared to the medical example. Moreover, the performance gains vary across the test datasets (see Fig. \ref{fig:med-results-bydataset} and Fig.~\ref{fig:all-results-bydataset}). In Appendix \ref{sec:Appendix-perf-gains}, we show through a toy example that FLAT delivers the highest performance gains when the pre-training tasks contain similar structural relationships between the variables as the downstream test tasks.

\section{Conclusions}

\textbf{Limitations \& Future work} The target network employs a fully connected graph between all columns, resulting in a time complexity of $O\left((N^{col}) ^2\right)$; therefore, operating on datasets with a large number of columns can be slow (Appendix \ref{Appendix-time}). We would also like to extend the FLAT architecture to multi-class learning with any number of classes as well as regression problems, e.g. by adding multiple classification heads. Finally, by masking out missing values, it becomes theoretically possible to work with incomplete datasets. Missing values in the meta datasets can be handled by omitting them from the sum in equation \ref{eq:d2v}, and missing target features can be handled by removing the corresponding node from the GAT. We leave these extensions for future research.

\textbf{Impact} We believe our work offers a valuable addition to the advancement of few-shot tabular learning. While traditional machine learning models often require vast amounts of data to train, FLAT enables meta-learning across datasets with heterogeneous feature spaces, reducing the need for large training datasets. This enhanced data efficiency can accelerate research and development in various domains. Some of the most common real-world scenarios with limited data are medical applications. Gathering extensive labeled patient data often proves challenging, particularly when dealing with rare conditions where imbalanced datasets are prevalent. For instance, FLAT presents a solution for the integration of datasets from several hospitals with potentially variable quantity and nature of recorded features in order to make improved predictions about patients’ health based on just a few labeled examples. 

\textbf{Summary} We present a new framework for few-shot learning on tabular datasets, an area that has been relatively underexplored despite its significance. Unlike most existing meta-learning methods that operate under the assumption of homogeneous feature spaces, our effectively handles diverse feature spaces, making it a novel solution in the meta-learning paradigm. To the best of our knowledge, the only other existing meta methods capable of addressing varying feature spaces are TabPFN and the model proposed by \cite{Iwata2020}, both of which, as demonstrated in our study, are outperformed by FLAT. Additionally, we highlight the importance of imbalanced learning in few-shot scenarios and demonstrate FLAT's effectiveness even on highly imbalanced datasets.


\clearpage
\bibliographystyle{unsrtnat}
\bibliography{literature}

\clearpage
\begin{appendix}
\onecolumn

\setcounter{table}{0}
\renewcommand{\thetable}{A\arabic{table}}
\setcounter{figure}{0}
\renewcommand{\thefigure}{A\arabic{figure}}

\section{Appendix}\label{sec:Appendix}
\subsection{Implementation details} \label{Appendix:details}
In this section, we provide a detailed description of the implementation of our model. 

To determine the hyperparameters for FLAT and the baselines, we performed tuning on a random subset of 40 out of the 118 datasets. From each selected dataset, 25\% of rows were randomly sampled to be used in validation. This collection of validation datasets is referred to as $\D_{val}$. Meta and target datasets were subsampled from the datasets in $\D_{val}$ in the same way as described in sec~\ref{sec:exp-evaluation}. This procedure ensured that all models' parameters were tuned on the same data.  Hyperparameter tuning on $\D_{val}$ was performed only once for each model and the selected parameters were used for all experiments. Tuning was performed at $N^{meta}=10$.

\subsubsection{FLAT}

\textbf{Dataset encoder $\mathcal{F}$} We base our implementation on the original Dataset2vec \citep{jomaa2021dataset2vec}. $f_1$ and $f_3$ are residual MLPs, each 4 sequential MLP blocks with skip connections between each intermediate layer. $f_2$ is a 2-layer MLP. The MLPs have hidden size 64 and output size 64 for the dataset embedding $\mathbf{e}$. ReLU activation functions are used for the entire model. 

\textbf{Column encoder $\mathcal{G}$} Our column encoder $\mathcal{G}$ is a 2-layer MLP with hidden dimension 64 and output dimension 15, which when concatenated with the column value gives a 16-dimensional vector as inputs to the target network $\Phi$. We initialize the output biases of this layer to 0 at the start of training. 

\textbf{Weight decoder $\mathcal{H}$} The weight generators $h_l$ are a series of linear MLPs with no bias terms. L2 weight normalization is applied on all generated weights with a learnable weight norm, one learnable norm is used for each GAT parameter (shared across GAT layers) and one for the final linear layer. We initialize the norms by training a model with initial norm 1, recording the final norm at the end of training and using this value as the new initialization for all training runs. 

\textbf{Target network $\Phi$} The target network, implemented as a GAT, has 2 heads, 2 layers, a hidden dimension of 128, and an output dimension of 16. We use a modified GAT implementation from PyTorch Geometric \citep{TorchGeometric} which allows for weight generation. The final classification layer is a single layer with an output size 2. A softmax layer is used for classification probabilities. 

\textbf{Optimization} Our model is trained using the AdamW \citep{AdamW} optimiser with lr=5e-4, eps=3-4, weight\_decay=1e-4. We train with batch size 3 for 62000 steps, taking around 11 minutes per model on a Ryzen 5800X3D CPU, depending on the dataset split used for training. 

\textbf{FLATadapt} Throughout this paper, FLATadapt uses the exact same already-trained FLAT models. FLATadapt uses 5 steps of gradient descent on $D^{meta}$ using the Adam optimizer \citep{Adam}. Column embeddings use lr=1e-3, and weight embeddings use lr=7.5e-2, all other parameters are AdamW defaults. Note that a higher learning rate is needed for the weight embedding. Only the dataset and column embeddings are changed in this process. FLATadapt only changes the inference process and not the training process. 

\subsubsection{Baselines}
The baselines used are based on existing / official implementations. 
Logistic regression, K-nearest neighbors, support vector classifier, and random forest use the scikit-learn implementation \citep{scikit-learn}. CatBoost \citep{CatBoost} used the Python implementation at \href{https://github.com/catboost/catboost/releases/tag/v1.1.1.}{https://github.com/catboost/catboost/releases/tag/v1.1.1.} \citep{arik2021tabnet} is based on the implementation at \href{https://github.com/dreamquark-ai/tabnet/releases/tag/v4.0.}{https://github.com/dreamquark-ai/tabnet/releases/tag/v4.0.} FT-Transformer \citep{FTtransformer} uses the implementation at \href{https://github.com/lucidrains/tab-transformer-pytorch/releases/tag/0.2.5.}{https://github.com/lucidrains/tab-transformer-pytorch/releases/tag/0.2.5.} 

Our STUNT implementation is modified based on the official implementation at \href{https://github.com/jaehyun513/STUNT}{https://github.com/jaehyun513/STUNT} \citep{nam2023stunt}. The original implementation assumes a very large unlabeled dataset but our unlabeled dataset, $D^{meta}$, is small. STUNT performs pre-training by using a random subset of columns to generate targets which fails if multiple columns are identical (it may not be possible to generate unique, balanced pseudo-labels from $D^{meta}$). This is more likely in our small unlabeled dataset. Therefore, we allow for reducing the number of shots during training. Furthermore, the use of a very small unlabeled dataset results in overfitting if STUNT is trained for many iterations. In our validation testing, we found a very low number (5) of training steps performed best. 

For each of the baselines (except logistic regression), we performed extensive manual parameter tuning on the validation data until we could no longer improve performance. Since our validation dataset is relatively large and we randomly sample rows and columns which acts as data augmentation, we are confident the parameters are not over-fit. To validate, we compare our tuned baselines to default baselines in Table \ref{tab:baselines} on a different random dataset collection to what was used for tuning. Note logistic regression and TabPFN have no tunable parameters and STUNT and TabNet do not have suitable default hyperparameters. Our tuned baselines are within error or better than the default baselines. 

\begin{table}[h]
    \centering
    \caption{Accuracy (\%) comparison between our tuned baselines vs default parameters with $N^{col} = 10$. Sampling errors are ± 0.25\%}
    {\small \include{figures/baseline_tuning}}
    \label{tab:baselines}
\end{table}

\subsection{Details of the main experiments}\label{Appendix-exp-details}

This subsection includes the remaining details of the experimental procedure used to report the results from sections \ref{sec:medical-example} and \ref{sec:generalist-exp}. First, we outline the details common for both the medical example (sec.~\ref{sec:medical-example}) and the general experiments (sec.~\ref{sec:generalist-exp}).

To create the training, $\mathcal{D}_{train}$, and testing, $\mathcal{D}_{test}$, collections of datasets we split the available datasets (29 for the medical example, 118 for the generalized scenario) into $N$ folds. We loop through all $N$ folds and use each fold as the testing collection once, while the remaining $N-1$ form the training collection. In this way, no samples used to pre-train FLAT belong to the same dataset as used during testing, ensuring a fair comparison against non-meta baselines fitted on just a few samples from $D^{meta}$ of each task. If a dataset is too small for a given $N^{meta}$, it is excluded from the training/testing collection. The meta training tasks are generated with a randomized sampling procedure including uniform sampling of the datasets from $\mathcal{D}_{train}$, binomial subsampling of $N^{meta} + N^{target}$ rows, and uniform sampling of columns. For testing, to ensure the reproducibility of the results and a fair comparison between the models, we sample 200 tasks per each dataset; these tasks are fixed for all models throughout all testing runs.
The errors reported in the tables are the standard deviation of predictions for each model, averaged over all $N$ testing folds. The errors for FLAT and FLATadapt are additionally averaged over several random initial seeds. 
The variance of the results comes from two factors: 1) the random sampling of testing tasks, which are the same for all models, 2) the model-specific variance for a given task. Since we evaluate all of our models on the exact same tasks, the differences in model performances have a lower variance than what the error bars indicate.

\textbf{Illustrative example: medical datasets} For the results presented in Table~\ref{tab:med-results}, FLAT was trained using meta and target datasets with 10 rows each ($N^{meta} = N^{target} = 10$) in order to demonstrate that FLAT can be used with different $N^{meta}$ during training and testing. The results for FLAT are averaged over 3 initial random seeds. We employed the $N$-fold validation strategy with $N=10$.

\textbf{Training a generalist few-shot learner} For the results in Table~\ref{tab:GenAll}, FLAT was trained on the same number of meta rows, $N^{meta}$ as during testing, with the exception of $N^{meta}=1$ and $N^{meta}=3$ where FLAT was trained with $N^{meta}=5$. $N^{target}$ was set to 15 during training and 5 for testing. Results for FLAT are averaged over 5 initial random seeds. We employed the $N$-fold validation with $N=4$.

\subsubsection{Accuracy per dataset for the main experiments}
\begin{figure}
    \centering
    \caption{Accuracy ($\%$) of FLAT vs. baseline models for the medical datasets. Evaluated on task datasets with $N^{meta}=5$. Columns (models) ordered by average model ranks. Rows (data sets) ordered by relative advantage of FLAT(adapt) vs. the best-performing baseline.}
    \includegraphics[width=0.65\textwidth]{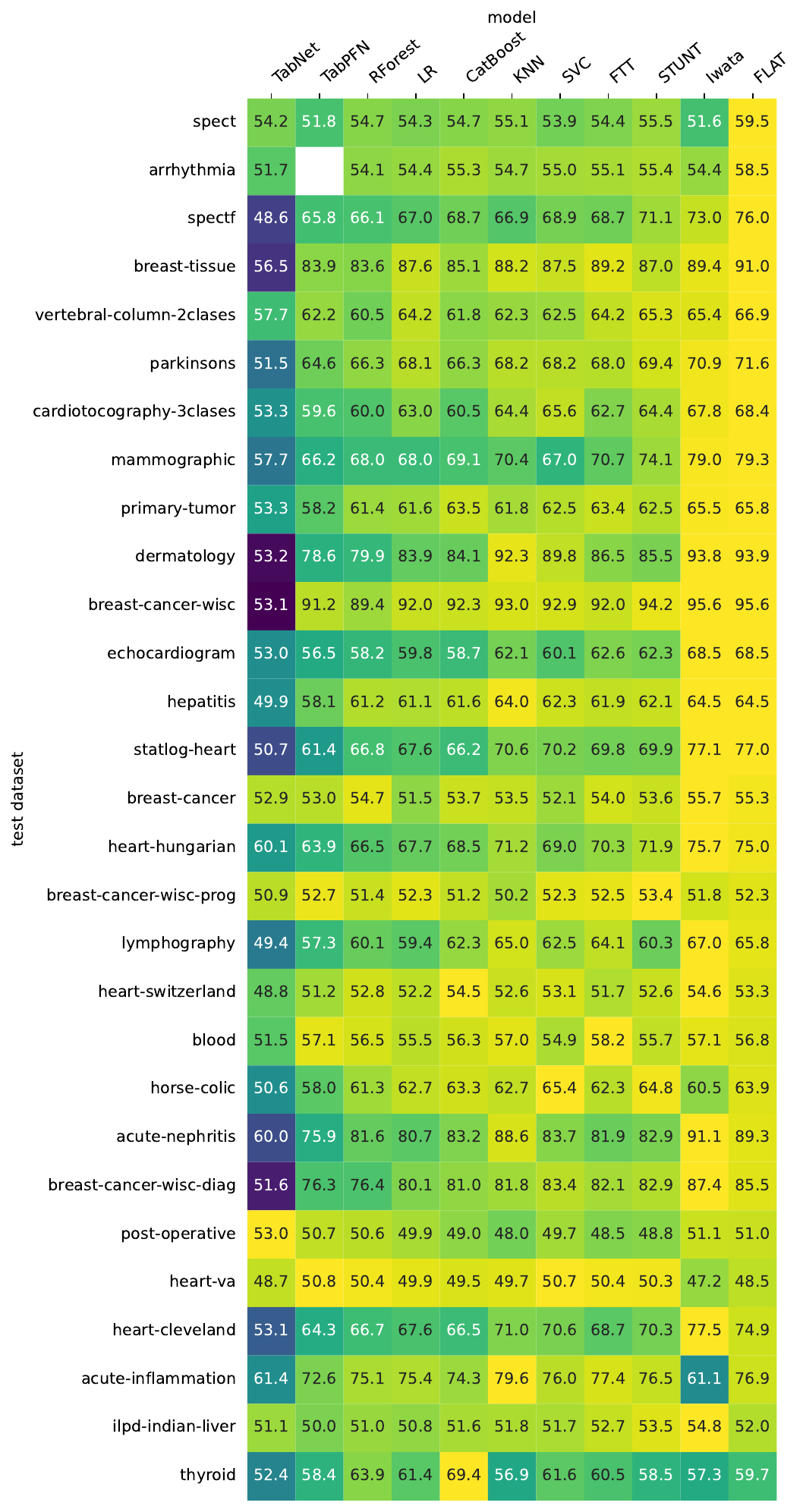}
    \label{fig:med-results-bydataset}
\end{figure}

\begin{figure}
    \centering
    \caption{Accuracy ($\%$) of FLAT vs. baseline models for all 118 datasets. Evaluated on task datasets with $N^{meta}=10$. Columns (models) ordered by average model ranks. Rows (data sets) ordered by relative advantage of FLAT(adapt) vs. the best-performing baseline.}
    \includegraphics[width=0.72\textwidth]{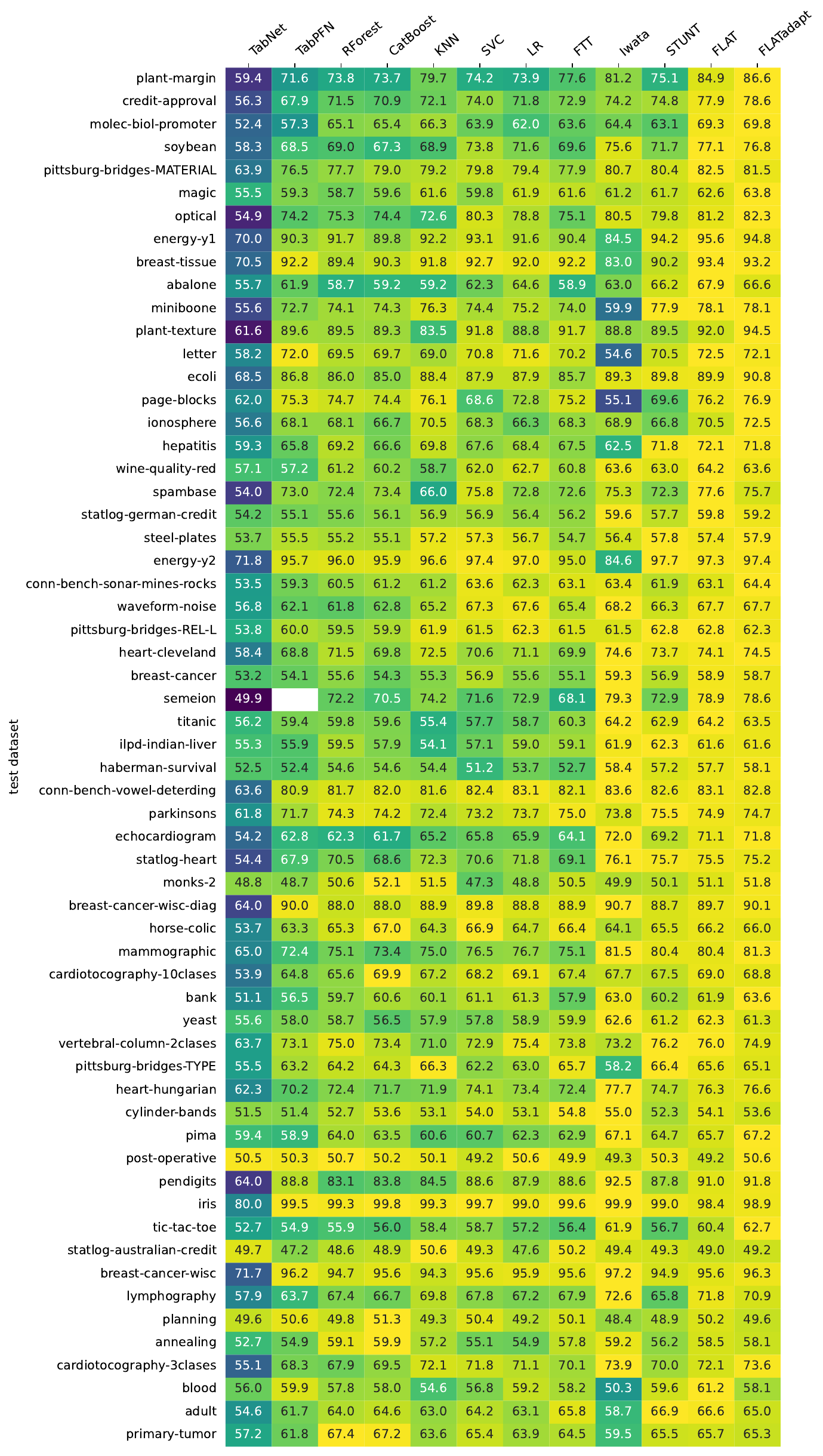}
    \label{fig:all-results-bydataset}
\end{figure}

Figures~\ref{fig:med-results-bydataset} and ~\ref{fig:all-results-bydataset} show the detailed results summarised in Tables~\ref{tab:med-results} and \ref{tab:GenAll} results respectively. Note that TabPFN is limited to datasets with at most 100 features. The accuracy of TabPFN on larger datasets, i.e. \textit{arrhythmia}, \textit{semeion}, \textit{hill-valley}, \textit{musk-1}, \textit{musk-2}, \textit{low-res-spect} are therefore missing.

\begin{figure}
\centering
\includegraphics[width=0.72\textwidth]{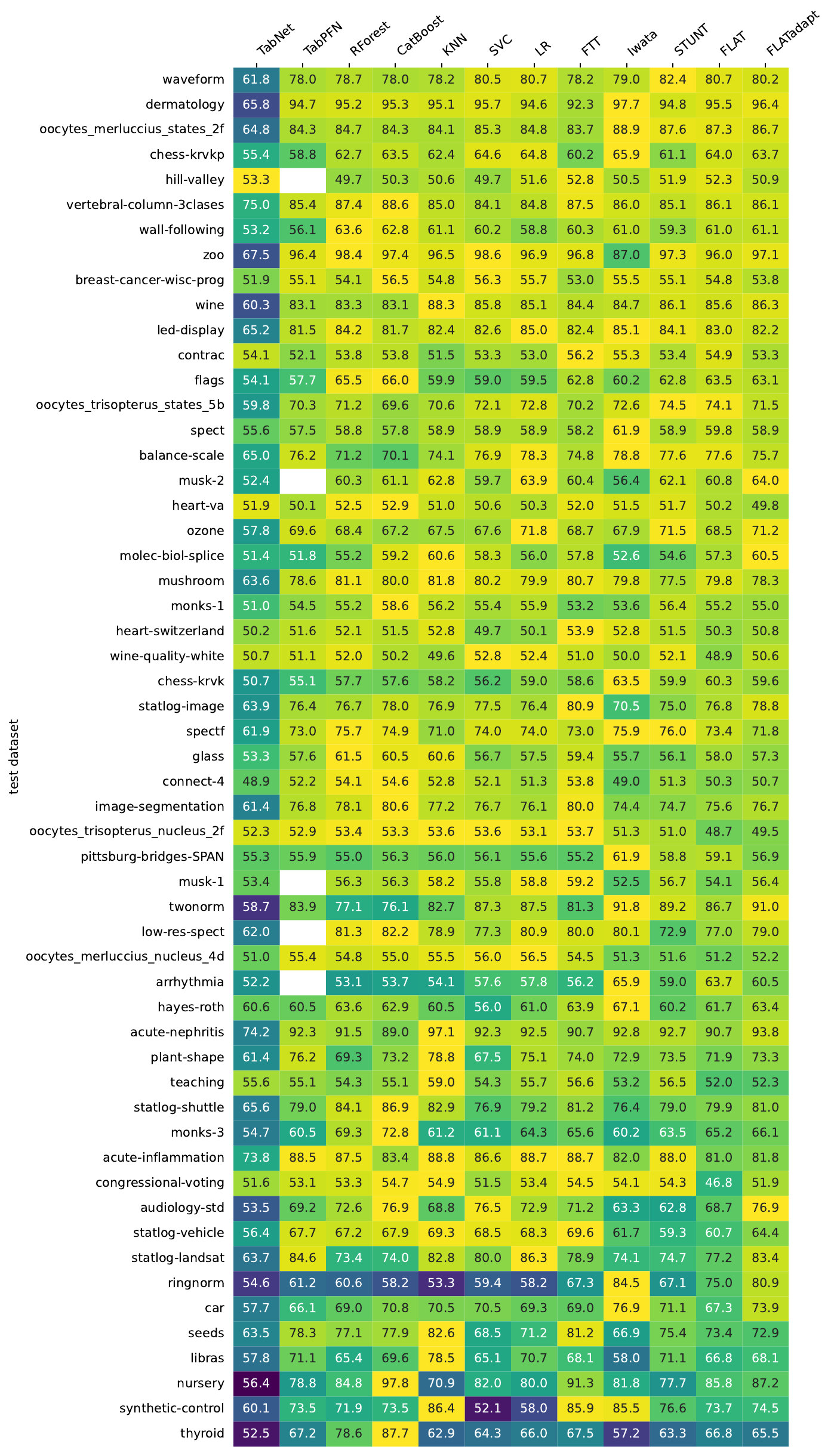}
\end{figure}

\subsection{Model interpretability}\label{sec:Appendix-interpretability}

\subsubsection{t-SNE embeddings}
\begin{figure}
    \centering
    \includegraphics[width=0.9\textwidth]{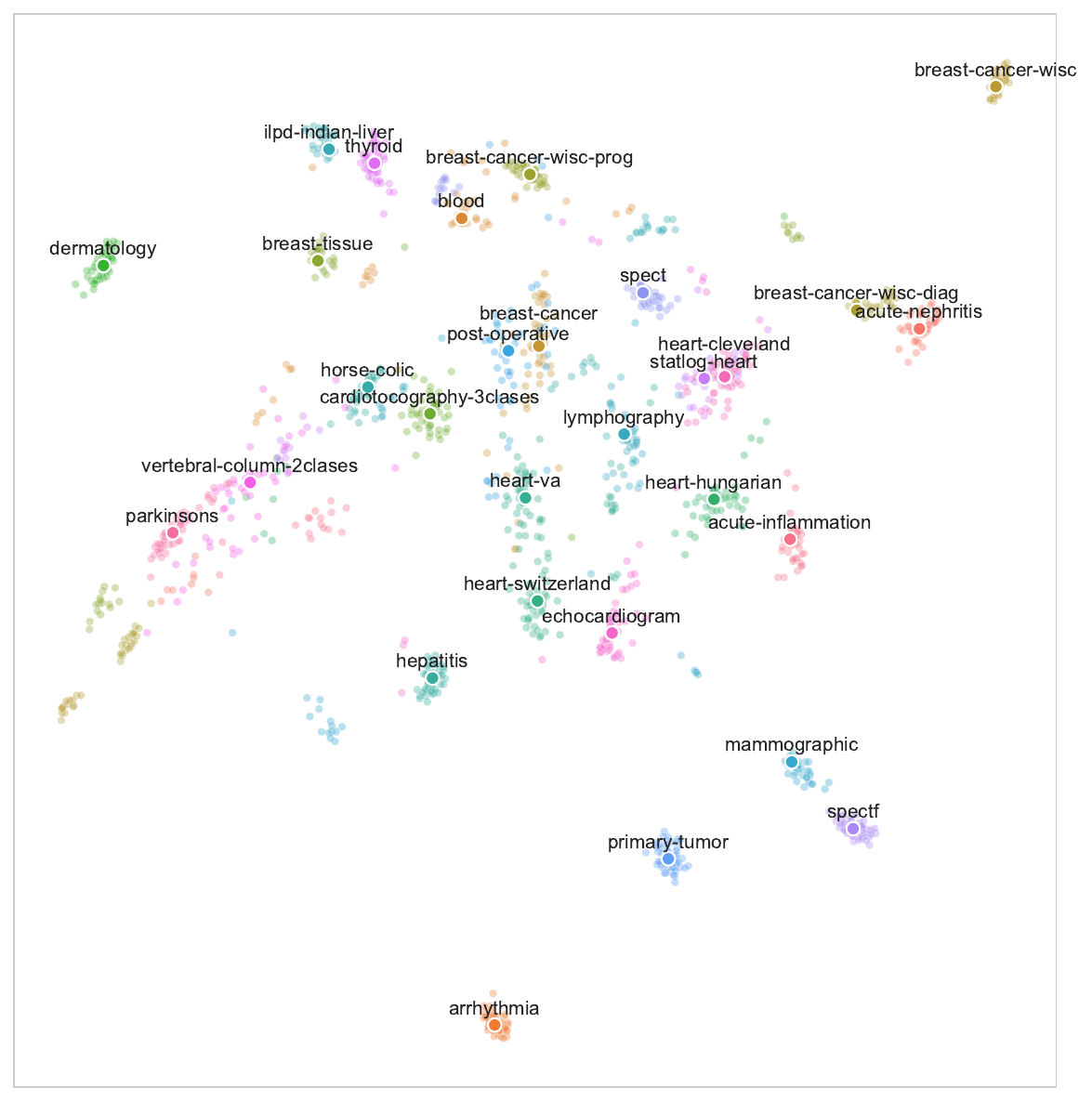}
    \caption{t-SNE plot of the medical datasets embeddings. The embeddings are generated for tasks coming from both $\D_{train}$ and $\D_{test}$ as defined by one of the 10 folds. In the above example $\D_{test} =$ \{\textit{echocardiogram, heart-cleveland, parkinsons}\}, the remaining datasets are included in the training collection. Lighter markers correspond to individual embeddings of each task. Bigger, darker markers with text annotations correspond to the geometric median computed for each dataset. The embeddings form clear clusters in agreement with their datasets.}
    \label{fig:tsne-med-data-embeddings}
\end{figure}

Fig.~\ref{fig:tsne-med-data-embeddings} depicts the same t-SNE embeddings as shown in Fig.~\ref{fig:medical-embeddings} from section~\ref{sec:medical-example} ($N^{meta} = 100$) with additional annotations of the centroids for each dataset, computed as the geometric median. The visualization of the embeddings enables us to gain further insight into which datasets are perceived as similar by the model. Specifically, the embeddings of the testing dataset \textit{heart-cleveland} are intermingled with the embeddings of the training dataset \textit{statlog-heart}, indicating a high degree of shared knowledge between the two datasets. This observation is particularly satisfying given that both datasets pertain to the cardiological conditions of patients, with the response variable representing the presence of heart disease. Furthermore, the \textit{echocardiogram} test dataset, which describes the survival of patients after a heart attack, is clustered close to the \textit{heart-switzerland} training dataset, which also deals with cardiological diseases. Finally, the \textit{parkinsons} test dataset is clustered next to the \textit{vertebral-column-2classes} training dataset. The \textit{parkinsons} dataset aims to discern healthy people from those with Parkinson's disease, while the response variable of the \textit{vertebral-column-2classes} corresponds to the presence of an abnormal vertebral column condition. According to \citet{leeparkinsons2018}, patients with Parkinson's disease are at a higher risk of developing osteoporotic vertebral compression fractures. These findings validate that FLAT can learn a highly expressive embedding space facilitating effective knowledge transfer for few-shot learning on tabular datasets.

\begin{figure}
    \centering
    \includegraphics[width=0.65\textwidth]{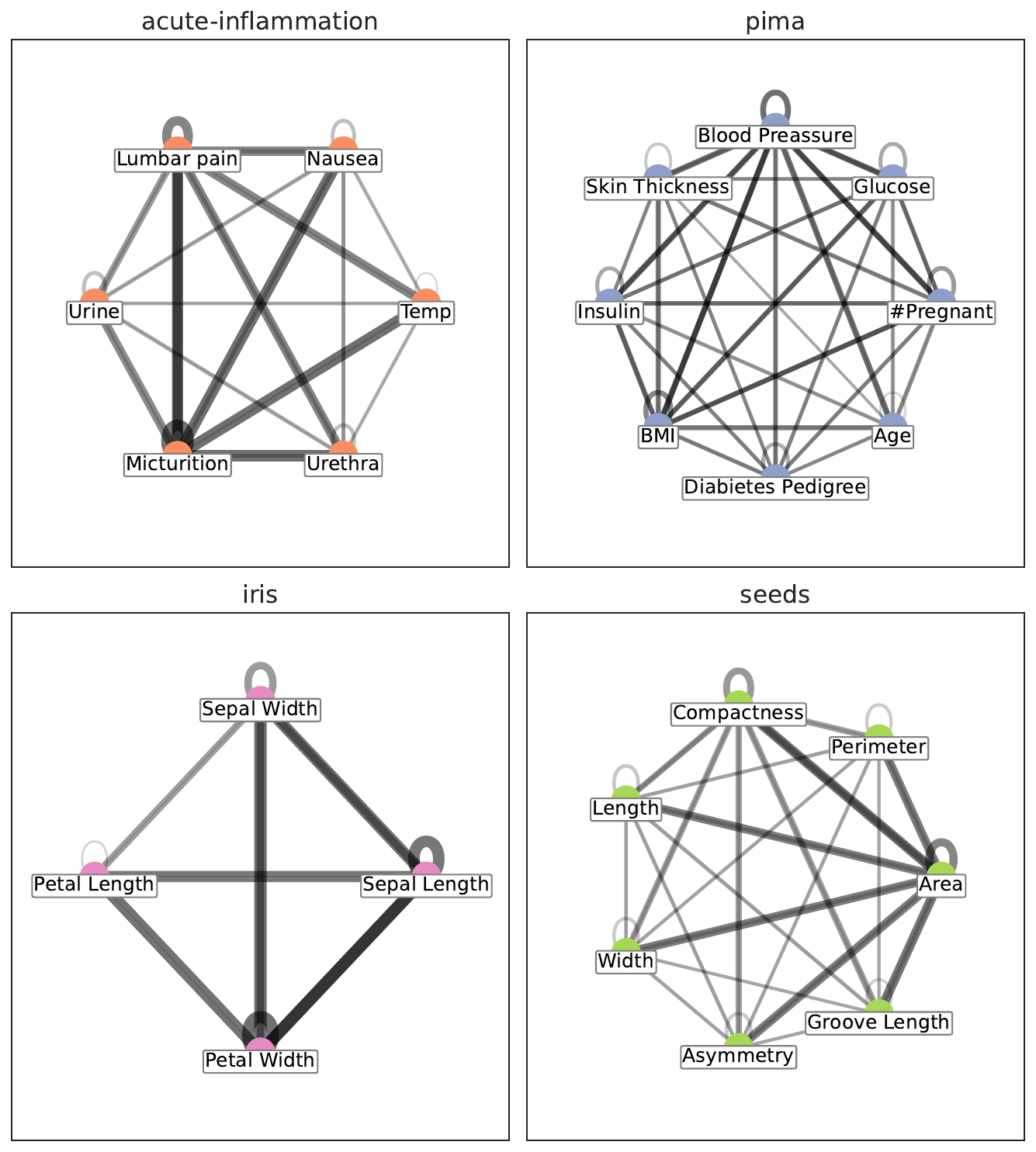}
    \caption{Plot of attention weights between nodes of the first layer of the GAT. Plots generated for 4 random subsamples of \textit{acute-inflammation, pima, iris}, and \textit{seeds} datasets.}
    \label{fig:gat_attention}
\end{figure}

\subsubsection{Attention maps}
The GAT produces attention maps which may be useful in determining what features the network focuses on. Between each pair of nodes, including itself, the attention weight determines how strongly to weigh each node's contributions, represented as $\alpha_{i,j}$ in Equation \ref{eq:GAT}. Nodes that have a higher weighting have more importance in the final result. In Fig.~\ref{fig:gat_attention}, we display the attention map for four random meta-datasets sampled from datasets that have their column names available. For instance, let's consider the \textit{acute-inflammation} dataset, which specifically focuses on urinary system diseases. In this dataset, we observe that the variable called \textit{Micturition} which indicates the presence of pain during urination, carries the highest weight within the meta-subsample. Another illustration is the \textit{seeds} dataset, which classifies different types of wheat. We can observe how the variable \textit{Area}, which measures the area of the kernels, carries the most weight.

\subsection{Additional experiments}

\subsubsection{Classic K-shot learning}\label{Appendix-classic-fewshot}
The research area of few-shot learning with imbalanced classes remains largely unexplored. This study expands on previous findings from the medical example presented in section \ref{sec:medical-example} by incorporating the standard definition of $K$-shot learning. Table~\ref{tab:kshot} presents a comparative analysis of the results for the FLAT model from section~\ref{sec:medical-example} tested on meta and target datasets containing an equal number of examples per class (equal \#labels) and tested using the randomized sampling method. The setting with an equal number of labels, where $N^{meta} = 2, 6, 10$, corresponds to the standard 1-, 3-, 5-shot learning definitions. The binomially sampled classes case is comparatively more challenging, which results in a decreased accuracy for the baseline models. The performance of FLAT remains the same under both sampling regimes and outperforms all baselines, except for the 5-shot case, where linear regression matches the performance of FLAT.

\begin{figure}[h]
    \centering
    \includegraphics[height=5cm]{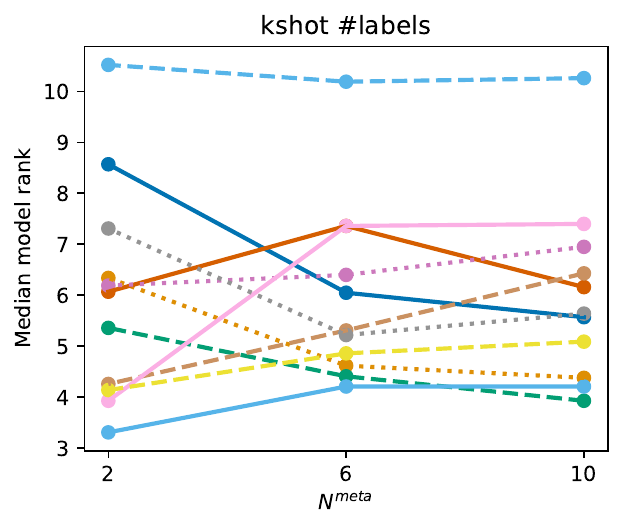}
    \quad
    \includegraphics[height=5cm]{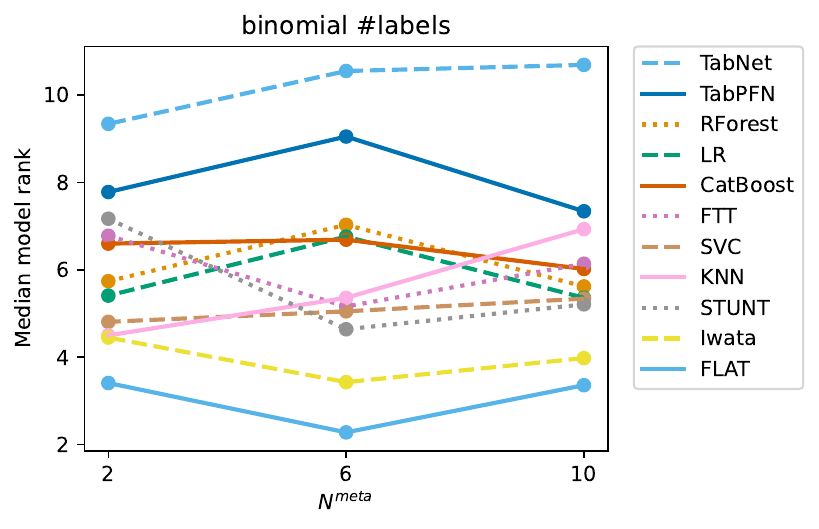}
    \caption{Median model ranks for equal sampling scheme (left) and binomial sampling (right).}
    \label{fig:kshot}
\end{figure}

\begin{table}
    \small
    \centering
    \caption{Comparison of imbalanced few-shot learning with standard $K$-shot learning on the 29 medical datasets. Accuracy of FLAT vs. the baselines when the number of examples per class is the same (equal \#labels), and when it is sampled from a binomial distribution (binomial \#labels).}
    {\small \include{figures/kshot_comparison}}
    \label{tab:kshot}
\end{table}

\subsubsection{Balance of the meta-dataset}
\begin{figure}
    \centering
    \includegraphics[width=0.45\textwidth]{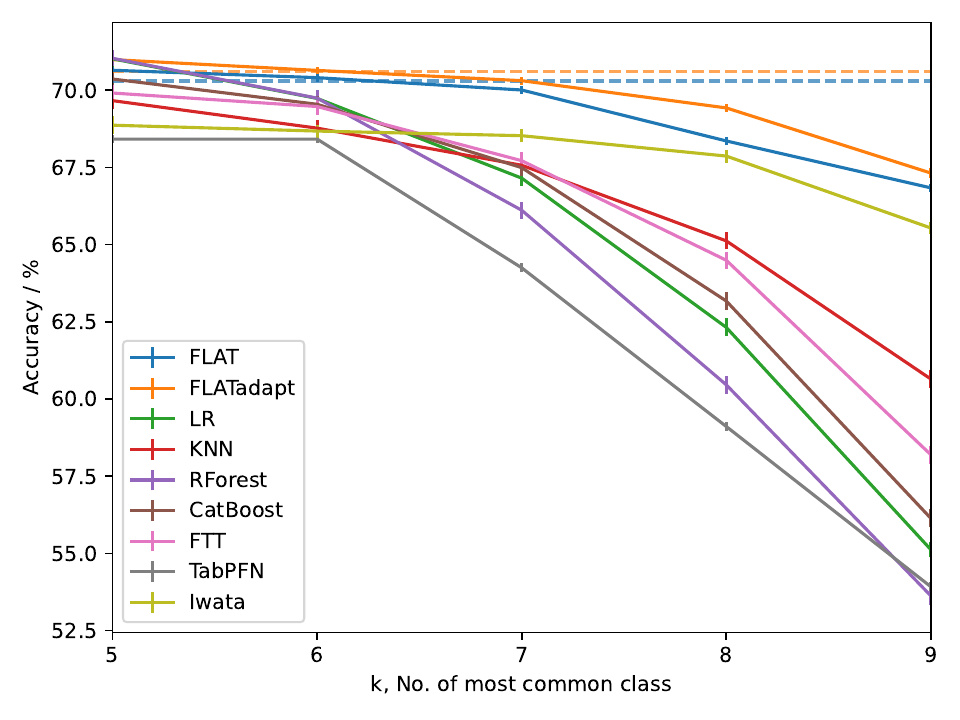}
    \caption{Model accuracy varying the balance of meta-dataset with total 10 rows, where $k=5$ is perfectly balanced and $k=1$ contains 9 instances of one class. Dotted lines show FLAT(adapt) accuracy when $k$ is binomially sampled. }
    \label{fig:vary_num_1s}
\end{figure}

This subsection explores the variability of FLAT predictions based on the balance of $D^{meta}$. We maintain a fixed size of 10 for $D^{meta}$ and sample $k$ positive samples per batch and the remainder of each batch with the opposite label. $k=5$ gives a balanced batch corresponding to the classic definition of 5-shot learning while increasing $k$ gives imbalanced batches. The binomial sampling scheme in the main paper is equivalent to re-sampling $k$ every batch with $k \sim \text{Bin(}p=0.5, n=N^{meta})$ (with the additional restriction of at least one example per class) which results in sampling balanced and imbalanced datasets, with the average dataset having an equal number of positive and negative labels. A plot of results as $k$ varies is shown in Figure \ref{fig:vary_num_1s}. The models used are the same as in the main paper, they are trained with $k \sim \text{Bin}(p=0.5, n=15)$ (excluding $k\in\{0, 15\}$). Retraining models with fixed $k$ during training would have likely given even better performance, though we did not try due to the number of models that would need to be trained and training on $n=15$ rows gave slightly improved performance vs $n=10$. FLATadapt outperforms FLAT and the other baselines, except at $k=5$ where linear regression and KNN match FLATadapt. All of the other models significantly drop in performance as the meta-dataset becomes more imbalanced, while FLAT(adapt) maintains strong performance. Note TabNet is excluded due to much lower accuracy than all the other baselines and very slow inference.

\subsubsection{When does FLAT result in large performance gains?}\label{sec:Appendix-perf-gains}

As observed, while FLAT delivers on average higher accuracy than the baseline models, the performance gains are not consistent across all testing datasets. These can vary anywhere between -2pp to +7pp. In this section we aim to further investigate conditions under which FLAT's pre-training is the most effective.

We hypothesise that some datasets used for testing may share few common characteristics with the training datasets, which could lead to inferior performance. If the model hasn't encountered certain feature-target relationships during training, its ability to leverage its prior knowledge during testing may be limited. We illustrate this with the following two experiments:

\textbf{Experiment 1} We identified 4 datasets with identical feature spaces. We visualized their correlation matrices and computed the pairwise Euclidean distances between them (Fig. R2). This analysis suggests that the heart-hungarian and heart-cleveland datasets exhibit high similarity, while heart-va is the most distinct. We conducted a leave-one-out testing procedure, where one dataset is used for testing and the remaining three are used for training. We expect that testing on heart-va would result in the lowest performance gains of FLAT, while testing on heart-cleveland or heart-hungarian, the highest. The results in Table R3 align with our expectations.

\textbf{Experiment 2} We further examined how the degree of similarity between the train and test datasets impacts performance. We selected heart-cleveland as the test dataset while the other 3 datasets were used for training. We sampled a subset of columns from the train and test datasets and varied the number of columns that overlap (i.e. columns in the intersection of the train and test datasets). Figure R3 shows how the performance gains of FLAT(adapt) versus baselines increases with the proportion of overlapping columns between training and test datasets.
Finally, we note that while FLAT may underperform on some datasets, no baseline consistently outperforms FLAT.

\begin{figure}[h]
    \centering
    \subfloat{\includegraphics[width=0.65\textwidth]{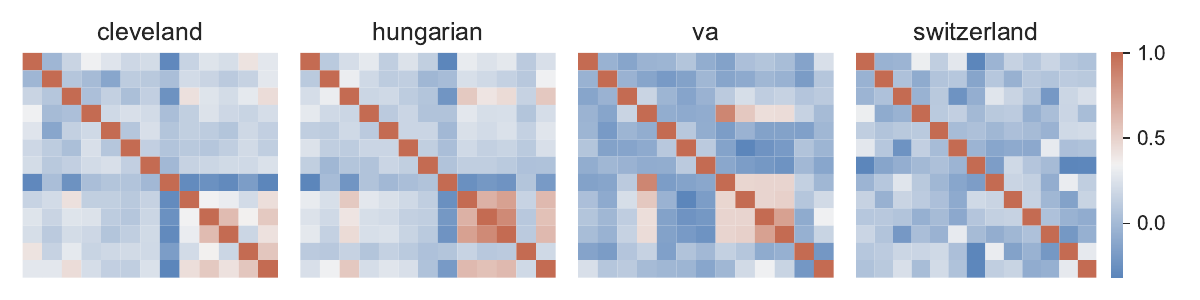}}%
    \qquad
    \subfloat{\includegraphics[width=0.25\textwidth]{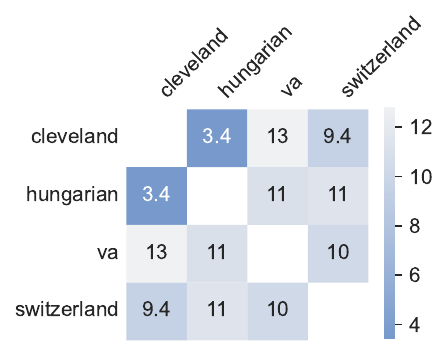}}%
    \caption{Correlation structure relationships between the four datasets used in experiment 1: \textit{heart-cleveland}, \textit{heart-hungarian}, \textit{heart-va} and \textit{heart-switzerland}. Left: Correlation matrices of the four datasets. Right: pairwise euclidean / frobenius distances between the correlation matrices.}%
    \label{fig:example}%
\end{figure}

\begin{table}[h]
  \centering
  \small
  \caption{Mean accuracy (\%) of FLAT and FLATadapt and mean performance gains (pp) over three baseline models. FLAT and FLATadapt exhibit the highest performance gain on \textit{heart-cleveland} and \textit{heart-hungarian} datasets. \textit{heart-va} does not benefit from FLAT's pretraining on the remaining datasets.}
  {\small \include{figures/experiment_1}}
  \label{}
\end{table}

\begin{figure}[h]
    \centering
    \includegraphics[height=3cm]{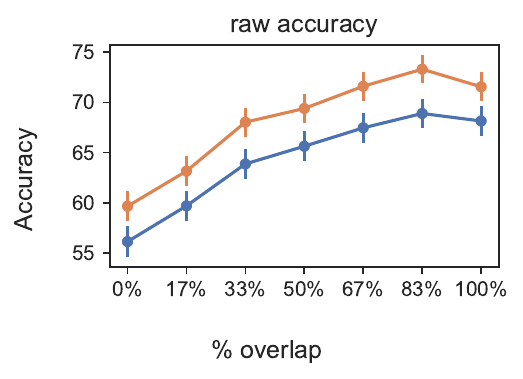}
    \includegraphics[height=3cm]{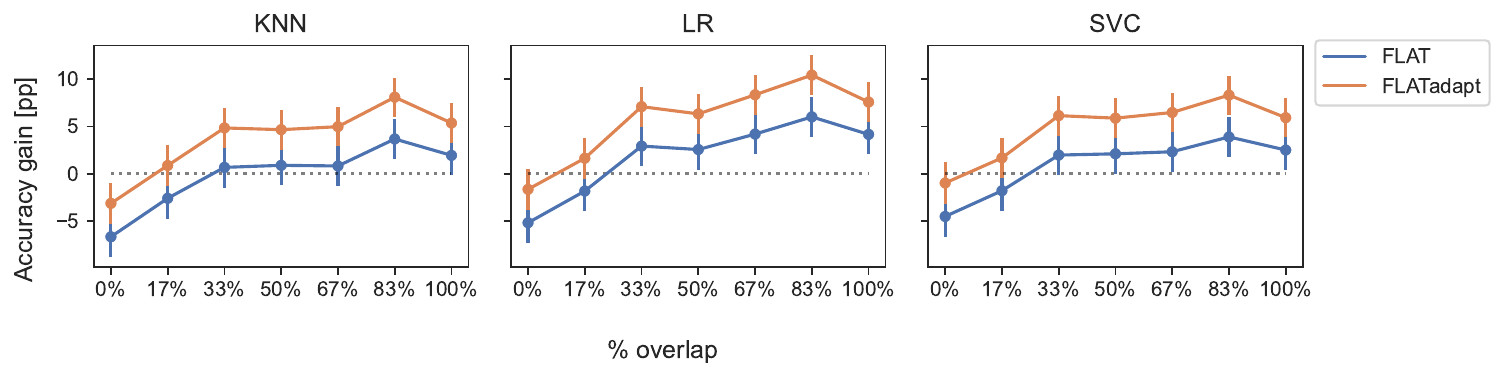}
    \caption{Accuracy of FLAT and FLATadapt and the performance gains over three baseline models. \%overlap is the proportion of columns which are common for training and testing datasets.}
    \label{fig:enter-label}
\end{figure}

\subsubsection{Multi-class classification}

Table \ref{tab:multiclass} presents the performance of FLAT(adapt) against the baselines on the 3-class classification tasks.

\begin{table}[h]
    \centering
    {\small \include{figures/multiclass}}
    \caption{3-class classification accuracy (\%) and succes no. on 65 UCI datasets.}
    \label{tab:multiclass}
\end{table}

\subsubsection{Predictions based on a single sample}\label{sec:Appendix-1-shot}

\begin{figure}
    \centering
    \includegraphics[width=0.6\textwidth]{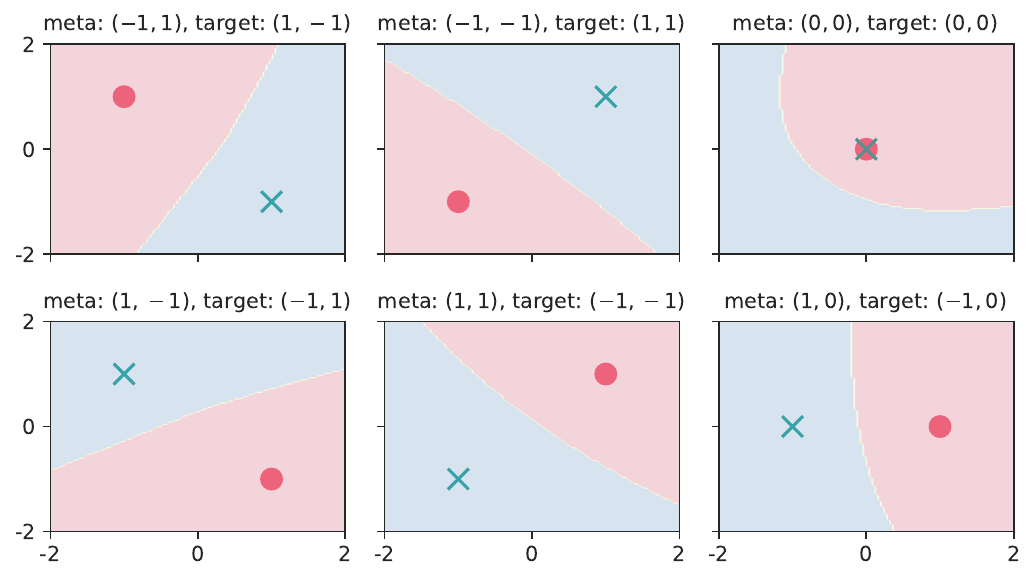}
    \caption{Decision boundaries for one-shot testing. Meta and target points represented as a red dot and a blue cross respectively.}
    \label{fig:one_shot}
\end{figure}

FLAT is able to make predictions with only a single labeled sample, whereas standard supervised models typically require at least one example from each class to perform inference. In Figure \ref{fig:one_shot},  we visualize classification boundaries obtained with one meta and one target sample. In our procedure, we jointly standardize features. As a result, identical features of a particular meta and target column are set to 0 and different features to $\pm 1$. In Fig. \ref{fig:one_shot} top right pane, when the meta and target values are the same for both coordinates, the same class is predicted for the target sample as the meta sample. In the remaining cases, i.e. where at least one feature differs, the opposite class is assigned. Also shown are the decision boundaries for if there were more than 1 target sample, allowing for feature values beyond $\{\pm1, 0\}$. Our model learns prior knowledge on how 'close' a target sample should be to the meta sample in order to be assigned the same class, by using standardization to fix the comparison scale.

\subsection{Inference time}\label{Appendix-time}

\begin{figure}
    \centering
    \includegraphics[width=0.4\textwidth]{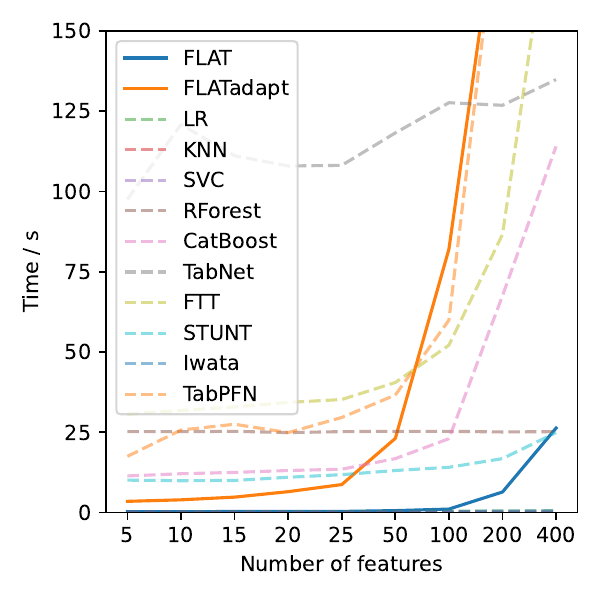}
    \caption{Inference time of different models vs. the number of features.}
    \label{fig:test-time}
\end{figure}

We perform additional inference time benchmarking, tracking the inference time versus the number of columns in $D^{meta}$ and $D^{target}$. We tested on up to 400 columns, which should cover many real-world dataset sizes. The results are presented in Fig.~\ref{fig:test-time}. We observe that the inference time for FLAT is lower than the majority of baselines. However, FLATadapt due to its additional extra adaptation steps is noticeably slower.

\subsubsection{In- vs. out-of-sample and -domain}

Tables~\ref{tab:med-results} and Fig.~\ref{fig:med-results-bydataset} present the results obtained from test datasets that were not used during the training process, all of which originate from the medical domain. Two questions may arise: 1. Does the performance of FLAT exhibit a significant decline when evaluated on unseen datasets, in comparison to the datasets used for training? In other words, does it suffer from overfitting to the training set? 2. Can a model trained on medical datasets be effectively applied to tasks derived from a different domain?

Table~\ref{tab:in-out-results} illustrates the average difference in accuracy between FLAT and the baseline models, where FLAT is trained on the medical collection of datasets as described in section~\ref{sec:medical-example}, and subsequently evaluated on the following: a) training datasets from the medical collection, b) test datasets from the medical collection, and c) test datasets from domains outside of medicine. Results were obtained on test tasks with $N^{meta}=5$. As anticipated, FLAT exhibits the highest relative advantage over the baseline models when tested on tasks generated from the datasets seen during training. Notably, when tested on unseen datasets from the medical domain, FLAT's performance decreases by a small amount (0.87pp). This indicates that FLAT does not suffer from overfitting to the training set and that FLAT is able to generalize to new, unseen tasks. Furthermore, FLAT, trained solely on the medical subset, demonstrates a comparably strong performance on unseen datasets from other domains. The way in which FLAT extracts and shares information between datasets is indeed invariant to the domain. Instead, what is fundamental for FLAT's inner workings are the structural relationships between the columns of the datasets. It is possible for a financial dataset, for instance, to exhibit structural similarities to a previously observed medical dataset, enabling knowledge sharing to occur regardless of the domain. However, as evident from Table~\ref{tab:in-out-results}, the performance improvement of FLAT is slightly higher when tested on the medical datasets, which aligns with our intuition that datasets from the same domain are more likely to share structural similarities.

\begin{table}
    \centering
    \small
    \caption{Average difference in accuracy between FLAT and the baseline models. Evaluated on the training datasets, and test datasets coming from both the same medical domain and from outside the domain.}
    {\small\include{figures/in-out-comparison}}
    \label{tab:in-out-results}
\end{table}

\subsubsection{Training on a varying number of columns}
We investigated how training on datasets with high or low $N^{col}$ affected performance on datasets with high or low $N^{col}$. We split our 118 datasets into 2 categories, datasets with $N^{col} > 40$ and datasets with $N^{col} \le 40$, denoted as $D_{large}$ and $D_{small}$. Within each split, train and test splits were constructed. A model was trained on each test split of $D_{large}$ and $D_{small}$ each model was tested on both test splits to see how the training $N^{col}$ affected test performance. Let $M_{large}$ and $M_{small}$ denote models trained on $D_{large}$ and $D_{small}$, respectively. The same methodology was employed as the rest of the paper; during training, the number of columns is uniformly sampled between 2 and the maximum number possible in a batch and at test time, all the columns are used. $M_{small}$ always trained on less than 40 columns in training while $M_{large}$ was trained on any number of columns up to the largest dataset. 
Results are shown in Table \ref{tab:n_col_train}. $M_{small}$ always perform much better than $M_{large}$. This is a surprising result, since we may expect $M_{small}$ to outperform on $D_{small}$ and $M_{large}$ to outperform on $D_{large}$. We suspect this is due to over-smoothing during training, since a large number of columns generates a very large fully connected graph in the target network, though we did not investigate further. FLATadapt improves the performance of the $M_{large}$. Note the model trained on $D_{small}$ generalized very well to $D_{large}$, despite never being trained on datasets with $N^{col} > 40$. We conclude that our model is able to generalize to $N^{col}$ unseen during training, provided it is trained on small $N^{col}$. Since the model generalizes so well to unseen $N^{col}$, it is likely not an important attribute in the latent embedding, $\mathbf{e}$. 

\begin{table}[h]
    \caption{Accuracy (\%) comparing models trained/tested on long/short datasets. Long datasets are datasets with more than 40 columns. Left shows FLAT, right shows FLATadapt with logistic regression (LR) shown for comparison.}
    \centering
    \small
    \begin{subtable}{0.4\textwidth}
        \centering
        \include{figures/long_short}
    \end{subtable}
    \begin{subtable}{0.4\textwidth}
        \centering
        \include{figures/long_short_maml}

    \end{subtable}
    \label{tab:n_col_train}
\end{table}


\end{appendix}
\end{document}

%% file: figures/medical_results.tex
\begin{tabular}{rcrrrrr}
\toprule
\multicolumn{1}{l}{} & \multicolumn{4}{c}{$N^{meta}$}    \\
\midrule
\textbf{model}       & \multicolumn{1}{c}{1*}         & \multicolumn{1}{c}{3} & \multicolumn{1}{c}{5} & \multicolumn{1}{c}{10} & \multicolumn{1}{c}{15} \\
\midrule
LR   & —                & 62.56 ± 0.28          & 64.47 ± 0.27          & 70.10 ± 0.26            & 72.69 ± 0.25           \\
KNN      & —            & 64.99 ± 0.27          & 65.99 ± 0.27          & 69.50 ± 0.26            & 70.58 ± 0.25           \\
SVC & —	&63.89 ± 0.27	&65.62 ± 0.27	&69.91 ± 0.26	&71.87 ± 0.25 \\
RForest	&—	&59.83 ± 0.28	&63.77 ± 0.28	&70.11 ± 0.26	&72.82 ± 0.25\\
CatBoost	&—	&62.86 ± 0.28	&64.90 ± 0.27	&69.89 ± 0.26	&72.44 ± 0.25\\
TabNet   & —            & 51.09 ± 0.29          & 53.10 ± 0.29           & 59.11 ± 0.29           & 61.75 ± 0.28           \\
FTT   & —     & 63.73 ± 0.27          & 65.67 ± 0.27          & 69.67 ± 0.26           & 72.17 ± 0.25           \\
STUNT	&—	&63.79 ± 0.28	& 66.02 ± 0.27	& 70.96 ± 0.26	& \textbf{72.87} ± 0.25 \\
TabPFN	&—	&59.24 ± 0.28	&62.51 ± 0.27	&69.23 ± 0.25	&72.00 ± 0.24\\
Iwata	&57.72 ± 0.64	&65.82 ± 0.60	&67.81 ± 0.59	&70.32 ± 0.57	&71.49 ± 0.56\\
FLAT    &\textbf{59.73} ± 0.18            & \textbf{66.54} ± 0.11 & \textbf{68.85} ± 0.10  & \textbf{71.83} ± 0.09  & \textbf{73.10} ± 0.11    \\
\bottomrule
\end{tabular}

%% file: figures/all_results.tex
\begin{tabular}{rcrrrrc}
\toprule
\multicolumn{1}{l}{} & \multicolumn{5}{c}{$N^{meta}$}  & \multicolumn{1}{l}{}  \\
\midrule
\textbf{model}      & \multicolumn{1}{c}{1} & \multicolumn{1}{c}{3} & \multicolumn{1}{c}{5} & \multicolumn{1}{c}{10} & \multicolumn{1}{c}{15} & \multicolumn{1}{c}{Time /s}\\
\midrule
LR                  &— & 60.37 ± 0.28 & 62.50 ± 0.28  & 68.62 ± 0.27 & 71.43 ± 0.26  & 0.42      \\
KNN                &— & 62.54 ± 0.28 & 64.19 ± 0.28  & 68.53 ± 0.27 & 70.54 ± 0.26  & 0.22    \\
SVC                 &— & 61.61 ± 0.28 & 63.54 ± 0.28  & 68.19 ± 0.26 & 70.28 ± 0.26  & \textbf{0.10}    \\
RForest         &— & 57.60 ± 0.29 & 60.83 ± 0.28  & 67.67 ± 0.27 & 70.95 ± 0.26  & 25.50    \\
CatBoost            &— & 60.53 ± 0.28 & 62.62 ± 0.28  & 68.67 ± 0.26 & \textbf{71.69} ± 0.26  & 12.22    \\
TabNet              &— & 51.08 ± 0.29 & 52.89 ± 0.29  & 58.00 ± 0.29 & 60.72 ± 0.29  & 108.42    \\
FTT    &— & 61.43 ± 0.28 & 63.73 ± 0.28  & 68.87 ± 0.26 & 69.94 ± 0.26  & 40.61     \\
STUNT               &— & 61.28 ± 0.28 & 63.64 ± 0.28  & 69.00 ± 0.26 & 70.99 ± 0.26  & 6.79 \\
TabPFN      &— & 57.06 ± 0.28 & 60.17 ± 0.28 & 66.98 ± 0.27 & 70.38 ± 0.26 & 24.9 \\
Iwata       &— &  62.48 ± 0.31 & 64.52 ± 0.31 & 68.04 ± 0.30 & 69.25 ± 0.30  & 0.27 \\
FLAT        &\textbf{58.83} ± 0.14 & \textbf{64.40} ± 0.13 & \textbf{66.40} ± 0.14  & 69.86 ± 0.12 & 71.50 ± 0.14  & 0.45    \\
FLATadapt   &\textbf{58.87} ± 0.13& \textbf{64.43} ± 0.10 & \textbf{66.52} ± 0.11  & \textbf{70.35} ± 0.12 & \textbf{71.89} ± 0.12 & 8.65     \\
\bottomrule
\end{tabular}
 

%% file: figures/baseline_tuning.tex
\begin{tabular}{rcrrrrrr}
\toprule
\textbf{model}       & KNN     & RForest & SVC     & CatBoost  & FTT & Iwata \\
\midrule
Base                 &  60.67  & 66.49       & 62.19   & 68.08  & 65.80  & 58.58        \\
Tuned                &  66.03  & 66.52       & 66.47   & 68.04  & 66.54  & 67.77    \\

\bottomrule
\end{tabular}

%% file: figures/kshot_comparison.tex
\begin{tabular}{r*{3}{c}r*{3}{c}}
\toprule
                                   & \multicolumn{7}{c}{$N^{meta}$}                                                                                                                                                                                               \\
\cmidrule(lr){2-8}
                                   & \multicolumn{3}{c}{\textbf{equal \#labels}}                                                        & \multicolumn{1}{c}{} & \multicolumn{3}{c}{\textbf{binomial \#labels}}                        \\
\cmidrule(lr){2-4}  \cmidrule(lr){5-8}
\textbf{model} & \multicolumn{1}{c}{\textbf{2}} & \multicolumn{1}{c}{\textbf{6}} & \multicolumn{1}{c}{\textbf{10}} & \multicolumn{1}{c}{} & \multicolumn{1}{c}{\textbf{2}} & \multicolumn{1}{c}{\textbf{6}} & \multicolumn{1}{c}{\textbf{10}} \\
\midrule
SVC           & 63.66 ± 0.27 & 69.23 ± 0.26 & 70.55 ± 0.25 & & 63.66 ± 0.27 & 66.68 ± 0.27 & 70.23 ± 0.26 \\
LR            & 63.30 ± 0.28 & 69.73 ± 0.26 & \textbf{71.75} ± 0.25 & & 63.36 ± 0.28 & 65.81 ± 0.27 & 70.39 ± 0.26 \\
CatBoost      & 62.75 ± 0.28 & 68.82 ± 0.26 & 70.91 ± 0.25 & & 62.36 ± 0.28 & 65.94 ± 0.27 & 70.45 ± 0.26 \\
RForest     & 62.70 ± 0.28 & \textbf{69.83} ± 0.26 & 71.97 ± 0.25 & & 63.06 ± 0.28 & 65.39 ± 0.27 & 70.64 ± 0.26 \\
KNN           & 63.99 ± 0.27 & 68.15 ± 0.26 & 69.89 ± 0.25 & &  64.05 ± 0.27 & 66.90 ± 0.27  & 69.51 ± 0.26 \\
TabNet        & 50.67 ± 0.29 & 56.02 ± 0.29 & 59.98 ± 0.28 & & 51.15 ± 0.29 & 54.37 ± 0.29 & 58.66 ± 0.29 \\
FTT           & 62.86 ± 0.28 & 68.52 ± 0.26 & 70.27 ± 0.26 & & 62.52 ± 0.28 & 66.59 ± 0.27 & 69.72 ± 0.26 \\
STUNT	& 62.32 ± 0.28	& 69.32 ± 0.26	& 71.22 ± 0.25 & & 62.62 ± 0.28	& 67.26 ± 0.27	& 70.93 ± 0.26 \\
TabPFN	&60.43 ± 0.27	&68.69 ± 0.25	&70.28 ± 0.24 & & 60.73 ± 0.27	&63.28 ± 0.26	&67.94 ± 0.25\\
Iwata	&64.05 ± 0.67	&68.97 ± 0.63	&70.32 ± 0.62	& &64.23 ± 0.67	&68.38 ± 0.64	&70.84 ± 0.62\\
FLAT    &\textbf{64.69} ± 0.12	&\textbf{70.07} ± 0.10	&71.53 ± 0.11  &   &\textbf{64.91} ± 0.11	&\textbf{69.88} ± 0.11	&\textbf{71.99} ± 0.10  \\
\bottomrule
\end{tabular}

%% file: figures/experiment_1.tex
\begin{tabular}{lccccccccc}
\toprule
& \multicolumn{2}{c}{raw accuracy}                         & \multicolumn{6}{c}{performance gains over the baselines}                                                                                      \\
& \multicolumn{2}{c}{-}                                    & \multicolumn{2}{c}{LR}                                   & \multicolumn{2}{c}{KNN}                                  & \multicolumn{2}{c}{SVC} \\
& \multicolumn{1}{c}{FLAT} & \multicolumn{1}{c}{FLATadapt} & \multicolumn{1}{c}{FLAT} & \multicolumn{1}{c}{FLATadapt} & \multicolumn{1}{c}{FLAT} & \multicolumn{1}{c}{FLATadapt} & FLAT     & FLATadapt    \\
\midrule
cleveland   & 76.37                    & 76.27                         & 5.17                     & 5.27                          & 3.77                     & 3.87                          & 5.67     & 5.77         \\
hungarian   & 77.83                    & 77.77                         & 4.37                     & 4.43                          & 5.87                     & 5.93                          & 3.67     & 3.73         \\
switzerland & 53.30                    & 52.30                         & 2.20                     & 3.20                          & -0.50                    & 0.50                          & 2.60     & 3.60         \\
va          & 51.00                    & 49.77                         & -0.53                    & 0.70                          & -1.23                    & 0.00                          & -0.83    & 0.40 \\
\bottomrule
\end{tabular}

%% file: figures/multiclass.tex
\begin{tabular}{rcrrrrc}
\toprule
\multicolumn{1}{l}{} & \multicolumn{4}{c}{$N^{meta}$}    \\
\midrule
\textbf{model}       & \multicolumn{1}{c}{3} & \multicolumn{1}{c}{5} & \multicolumn{1}{c}{10} & \multicolumn{1}{c}{15}\\
\midrule
LR                   & 47.48 ± 0.39  & 54.48 ± 0.38  & 63.89 ± 0.35  & 68.57 ± 0.34     \\
KNN                & 48.90 ± 0.39 & 56.04 ± 0.38  & 63.98 ± 0.35  & 67.62 ± 0.34     \\
SVC                  & 48.44 ± 0.39 & 55.52 ± 0.38  & 63.68 ± 0.35  & 67.72 ± 0.34   \\
RForest          & 44.48 ± 0.39 & 52.42 ± 0.38   & 62.69 ± 0.36 & \textbf{68.59} ± 0.34     \\
CatBoost            & 47.42 ± 0.39 & 53.59 ± 0.38  & 63.51 ± 0.36 & \textbf{69.02} ± 0.34     \\
FTT     & 47.55 ± 0.39 & 54.54 ± 0.38  & 62.62 ± 0.36 & 67.07 ± 0.35      \\
STUNT              & 51.93 ± 0.39 & 56.11 ± 0.37  &63.78 ± 0.35 & 67.47 ± 0.34   \\
TabPFN       & 44.50 ± 0.39 & 50.38 ± 0.38 & 60.65 ± 0.36 & 66.48 ± 0.34 \\
Iwata        &  51.11 ± 0.43 & 55.11 ± 0.42 & 59.71 ± 0.41 & 62.09 ± 0.36  \\
FLAT         & \textbf{54.49} ± 0.35  & 58.77 ± 0.40 & 64.71 ± 0.38  & 67.31 ± 0.44 &   \\
FLATadapt    & \textbf{55.03} ± 0.32& \textbf{59.57} ± 0.35 & \textbf{65.61} ± 0.32  & \textbf{68.55} ± 0.33      \\

\bottomrule
\end{tabular}

%% file: figures/in-out-comparison.tex
\begin{tabular}{rrrr}
\toprule
\textbf{dataset split}                          & \multicolumn{2}{c}{\textbf{test}}                                  & \multicolumn{1}{c}{\textbf{train}} \\
\cmidrule(l{5pt}r{5pt}){2-3} \cmidrule(l{5pt}r{5pt}){4-4}
\textbf{medical}                                & \multicolumn{1}{c}{\xmark} & \multicolumn{1}{c}{\cmark} & \multicolumn{1}{c}{\cmark}   \\
\midrule
CatBoost                                        & 3.73                            & 3.95                             & 4.82                               \\
FTT                                   & 2.56                            & 3.18                             & 4.05                               \\
KNN                                             & 2.01                            & 2.87                             & 3.74                               \\
LR                                              & 3.74                            & 4.38                             & 5.25                               \\
RForest                                         & 5.10                             & 5.09                             & 5.96                               \\
STUNT                                           & 2.73                            & 2.83                             & 3.70                                \\
SVC                                             & 2.73                            & 3.23                             & 4.10                                \\
TabNet                                          & 12.72                           & 15.75                            & 16.63                              \\
\midrule
\multicolumn{1}{l}{\textbf{average difference}} & 4.42                            & 5.16                             & 6.03        \\
\bottomrule
\end{tabular}

%% file: figures/long_short.tex
\begin{tabular}{ccc}

\toprule
              &\multicolumn{2}{c}{\textbf{test}} \\
\textbf{train}      & short     & long \\
\midrule
short                 &  72.86  & 71.75     \\
long                &  59.92  & 63.42      \\
LR                   &  71.02  &  68.41     \\
\bottomrule
\end{tabular}

%% file: figures/long_short_maml.tex
\begin{tabular}{ccc}

\toprule
              &\multicolumn{2}{c}{\textbf{test}} \\
 \textbf{train}        & short     & long \\
\midrule
short                 &  72.64  & 72.97     \\
long                  &  69.52  & 66.26      \\
LR                   &  71.02  &  68.41     \\
\bottomrule
\end{tabular}